\documentclass[pdflatex,sn-mathphys-num]{sn-jnl}

\usepackage{graphicx}%
\usepackage{multirow}%
\usepackage{amsmath,amssymb,amsfonts}%
\usepackage{amsthm}%
\usepackage{mathrsfs}%
\usepackage[title]{appendix}%
\usepackage{xcolor}%
\usepackage{textcomp}%
\usepackage{manyfoot}%
\usepackage{booktabs}%
\usepackage{algorithm}%
\usepackage{algorithmicx}%
\usepackage{algpseudocode}%
\usepackage{listings}%
\usepackage{array}

\theoremstyle{thmstyleone}%
\theoremstyle{thmstyletwo}%

\theoremstyle{thmstylethree}%

\begin{document}

\title[NeuralMAG: Fast and Generalizable Micromagnetic Simulation with Deep Neural Nets]{NeuralMAG: Fast and Generalizable Micromagnetic Simulation with Deep Neural Nets}


\author[1]{\fnm{Yunqi} \sur{Cai}}\email{yunqicai2017@gmail.com}

\author*[2]{\fnm{Jiangnan} \sur{Li}}\email{li-jn12@tsinghua.org.cn}
\equalcont{These authors contributed equally to this work.}

\author*[3]{\fnm{Dong} \sur{Wang}}\email{wangdong99@mails.tsinghua.edu.cn}
\equalcont{These authors contributed equally to this work.}

\affil*[1]{\orgname{Faculty of Information Engineering and Automation, Kunming University of Science and Technology}, \orgaddress{\city{Kunming}, \postcode{650504}, \country{China}}}

\affil*[2]{\orgname{Faculty of Materials Science and Engineering, Kunming University of Science and Technology}, \orgaddress{\city{Kunming}, \postcode{650093}, \country{China}}}

\affil*[3]{\orgname{BNRist at Tsinghua University}, \orgaddress{\city{Beijing}, \postcode{100084}, \country{China}}}



\abstract{
Micromagnetics has made significant strides, particularly due to its wide-ranging applications in magnetic storage design and the recent exciting advancements in spintronics research. Numerical simulation is a cornerstone of micromagnetics research, relying on first-principle rules to compute the dynamic evolution of micromagnetic systems based on the renowned LLG equation, named after Landau, Lifshitz, and Gilbert. However, simulations are often hindered by their slow speed, primarily due to the global convolution required to compute the demagnetizing field, which involves full interaction among any two units in the sample. Although Fast-Fourier transformation (FFT) calculations reduce the computational complexity to O(NlogN), it remains impractical for large-scale simulations. In this paper, we introduce NeuralMAG, a deep learning approach to micromagnetic simulation. Our innovative approach follows the LLG iterative framework but accelerates demagnetizing field computation through the employment of a U-shaped neural network (Unet). The Unet architecture comprises an encoder that extracts aggregated spins at various scales and learns the local interaction at each scale, followed by a decoder that accumulates the local interactions at different scales to approximate the global convolution. This divide-and-accumulate scheme achieves a time complexity of O(N), significantly enhancing the speed and feasibility of large-scale simulations. Unlike existing neural methods, NeuralMAG concentrates on the core computation rather than an end-to-end approximation for a specific task, making it inherently generalizable. To validate the new approach, we trained a single model and evaluated it on two micromagnetics tasks with various sample sizes, shapes, and material settings: (1) basic LLG dynamic evolution, and (2) MH curve estimation. The results show that the model maintains reasonable accuracy and is significantly faster than the conventional FFT-based method, achieving a sixfold speedup for large-size models. NeuralMAG has been published online and is available for users to download.
}

\maketitle

\section{Main}\label{sec1}

\begin{figure}[h]
    \centering
    \includegraphics[width=\textwidth]{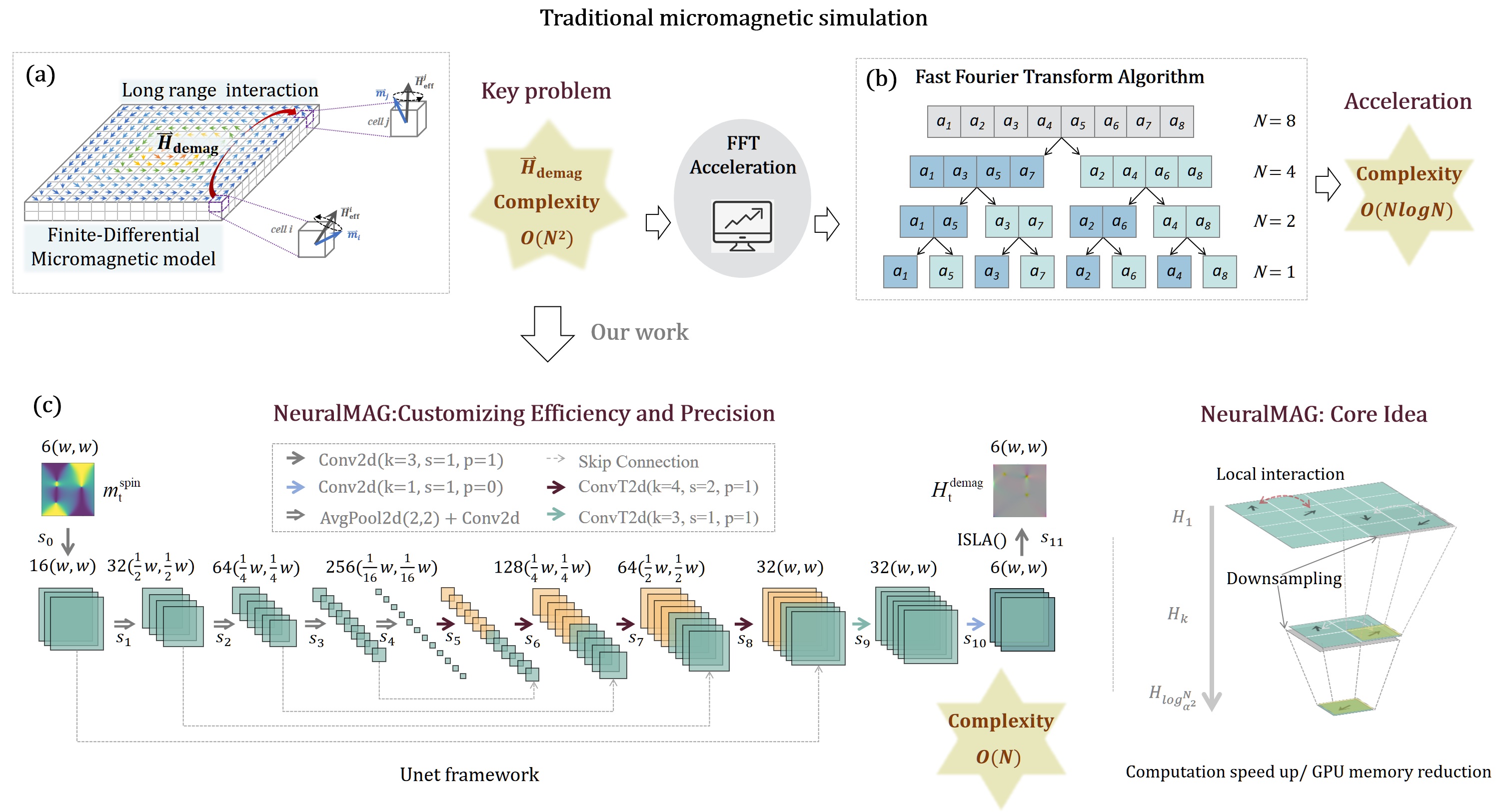}
    \caption{
    (a) Landau-Lifshitz-Gilbert (LLG) dynamics solved by the Finite Differential Method (FDM), focusing on the computational complexity caused by cell interactions. It highlights that the major issue with this framework is its large computational demand ($O(N^2)$), mainly due to the long-range cross-cell interaction.
    (b) The demagnetization equation (referenced in Eq. \ref{eq:demag}) is represented as a convolution of the magnetization vector \(\protect\overrightarrow{m}\) and the demagnetization tensor \(\protect\Omega\). By utilizing Fast Fourier Transform (FFT), this convolution is transformed into a spectral domain multiplication, significantly reducing the computational complexity to \(O(N\log N)\).
    (c) The NeuralMAG framework utilizes a Unet model to calculate  the demagnetizing field, the same role as FFT. This method accumulates local convolution outputs at varied granularity to approximate the global convolution between the magnetization vector \(\protect\overrightarrow{m}\) and the demagnetization tensor \(\Omega\). The core idea of this approach is depicted on the right side, where local cross-cell interaction is computed by a convolution layer, and a downsampling layer pools the neighbouring cells to form the next layer of cells with a larger scale. This convolution \& downampling operations continue untill the feature maps shrink to a set of $1 \times 1$ channel maps. Accumulating the cross-cell interactions at all levels of the hierarchy can lead to an approximation for the demagnetizing field with high accuracy, with a computational complexity $O(N)$ under mild conditions. Refer to the Discussion section for details.
    }
    \label{fig:frame}
\end{figure}

Born in the early 20th century to address the issues of magnetic domain and hysteresis, micromagnetics has evolved into the fundamental methodology for understanding the magnetic behavior of materials from a microscopic view \cite{landau1992theory,kittel1949physical}. The practicality of micromagnetics has earned itself popularity within the communities of magnetic storage and permanent magnets, alongside growing demands at the forefront of spintronics research \cite{victora2013simulation,zhu2007microwave,fert2013skyrmions,zhang2016antiferromagnetic,tehrani2000recent,sepehri2014micromagnetic,loewe2017grain,abert2019micromagnetics,dieny2020opportunities,barker2021review}. The theoretical framework, established by L. Landau and further developed by W. F. Brown \cite{landau1992theory,brown1959micromagnetics}, aims to account for various forms of magnetic energies and to determine the system's evolutionary path based on the energy landscape. This evolutionary process is captured by the renowned Landau-Lifshitz-Gilbert (LLG) dynamics \cite{gilbert2004phenomenological}, as illustrated in Fig.~\ref{fig:frame}(a) and detailed below:

\begin{equation}
\label{eq:llg}
\frac{\rm{d} \overrightarrow{m}}{\rm{d}t} = -\gamma \overrightarrow{m} \times \overrightarrow{H}_{eff} - \lambda \overrightarrow{m} \times (\overrightarrow{m} \times \overrightarrow{H}_{eff})
\end{equation}

\noindent where $\overrightarrow{m}$ is the magnetization vector, $\overrightarrow{H}_{eff}$ the effective field, $\gamma$ the electron gyromagnetic ratio, and $\lambda$ a phenomenological damping parameter. The effective field $\overrightarrow{H}_{eff}$ serves as a consolidated representation of magnetic energies, encompassing the external magnetic field $\overrightarrow{H}_{ext}$ derived from Zeeman energy, the anisotropy field $\overrightarrow{H}_{aniso}$ from magnetocrystalline energy, the demagnetizing field $\overrightarrow{H}_{demag}$ due to magnetostatic interaction, and the exchange field $\overrightarrow{H}_{exch}$ from the Heisenberg exchange interaction, as follows:

\begin{equation}
\label{eq:eff}
\overrightarrow{H}_{eff} = \overrightarrow{H}_{ext} + \overrightarrow{H}_{aniso} + \overrightarrow{H}_{demag} + \overrightarrow{H}_{exch}
\end{equation}

The above equation is typically solved using numerical computation methods, among which the finite-differential method (FDM) and the finite-element method (FEM) are most commonly employed. However, a significant challenge arises from the computation of the demagnetizing field $\overrightarrow{H}_{demag}$ (also referred to as the magnetostatic field), which is notoriously difficult. In the context of the FDM, for any specific cell denoted as $(i,j,k)$, the primary computation of $\overrightarrow{H}_{demag}(i,j,k)$ requires summing up the interactions between this cell and every other cell in the model, represented as $(l,m,n)$:

\begin{equation}
\label{eq:demag}
\overrightarrow{H}_{demag}(i,j,k) = M_s \sum_{lmn} \Omega(l-i, m-j, n-k) \cdot \overrightarrow{m}_{lmn}
\end{equation}

\noindent where $M_s$ represents the magnitude of magnetization, known as saturation magnetization, and $\Omega$ represents the location-invariant magnetostatic interaction tensor \cite{schabes1987magnetostatic}. Crucially, the strength of magnetostatic interactions adheres to a 1/r law, indicating that long-distance interactions might be sufficiently strong and so cannot be simply disregarded, resulting in a computational complexity of $O(N^2)$. This high complexity has emerged as a significant concern within micromagnetic studies, leading to simulations being limited to small-scale, simplified tasks, which notably curtails their utility in real-world applications. In 1988, Zhu et al. introduced Fast-Fourier transformation (FFT) calculations into the FDM scheme \cite{zhu1988micromagnetic}, by noticing that Eq.~\ref{eq:demag} is essentially a convolution between $\overrightarrow{m}$ and $\Omega$. This method dramatically reduced the complexity of magnetostatic calculations from $O(N^2)$ to $O(N\log(N))$, quickly elevating micromagnetic simulation into a powerful tool for designing hard-disk drivers (HDDs) \cite{zhu1988micromagnetic}. However, despite these improvements, FFT computations, as illustrated in Fig.~\ref{fig:frame}(b), still struggle to scale to large-size problems. For instance, the size of a feasibly modeled permanent magnet remains commonly below one micron, a scale even smaller than a single grain in actual materials\cite{zhao2022intrinsically,bjork2023explaining,tang2023unveiling}. The enhancement of computation efficiency is eternally required in micromagnetic research.

\begin{figure}[h]
    \centering
    \includegraphics[width=\textwidth]{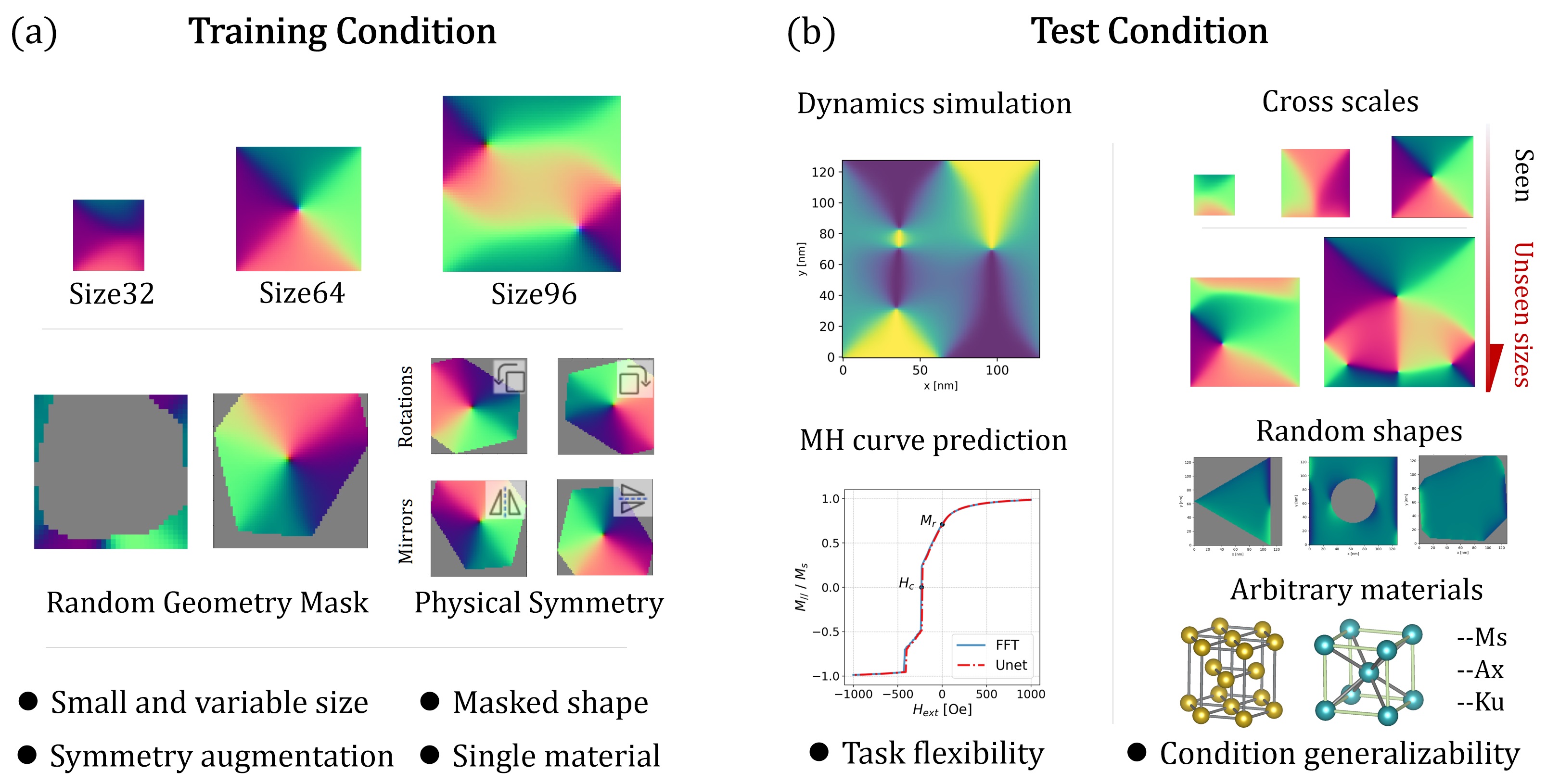}
    \caption{
    (a) Training condition for the U-Net model in the NeuralMAG framework. It involve using material samples of varying sizes, capped at a maximum size of 96 to maintain training efficiency. To enhance the model's robustness, samples are randomly masked. Furthermore, symmetric augmentation techniques are applied to ensure the model adheres to physical laws, reinforcing its ability to generalize across different micromagnetic scenarios. (b) Test condition of the U-Net model.  We evaluate the applicability of the model to diverse tasks, i.e., dynamics simulation and MH curve prediction in this study.  We also evaluate its generalizability by testing the performance of the model with  samples of different sizes, shapes, and materials.
    }
    \label{fig:CoreIdea}
\end{figure}

Recently, the rapid evolution of deep learning (DL) methods has brought a significant revolution in traditional computational physics\cite{zhang2018deep,wang2023scientific,baldi2014searching,li2022deep,karagiorgi2022machine,merchant2023scaling,karniadakis2021physics,schutt2019unifying,rao2023encoding,thiyagalingam2022scientific,shi2021towards,huerta2019enabling}. In micromagnetics, significant efforts have been devoted to training DL models as alternative tools for magnetostatic calculation. Khan et al. trained a convolutional neural network (CNN) to calculate the $\overrightarrow{H}_{demag}$ distribution in electromagnetic motors \cite{khan2019deep}. Kovacs et al. implemented a physics-informed neural network (PINN) to substitute the calculation of magnetostatic interaction in the FEM micromagnetic scheme \cite{kovacs2022magnetostatics}. The high accuracy of PINN on the $\mu MAG$ standard problem $\#$3 was later demonstrated by Schaffer et al \cite{schaffer2023physics}. Additionally, some research efforts focused on learning the direct time evolution of the magnetization state. For instance, Kovacs et al. trained a neural network to simulate the magnetic dynamic process \cite{kovacs2019learning} and demonstrated their approach on the $\mu MAG$ standard problem $\#$4. Chen et al. used neural ordinary differential equations (NODE) to reproduce the trajectory of magnetic skyrmions \cite{chen2022forecasting}. An acceleration factor exceeding 200 times was reported in comparison to traditional micromagnetic simulation. Despite these promising results, the reported DL models for micromagnetics were typically designed to solve specific problems, with little focus on generalization across various scales and configurations. Consequently, none of them can yet be considered a general computational tool for micromagnetic simulations.

In this study, we introduce NeuralMAG, a general and generalizable computation framework for micromagnetic simulation by integrating deep learning methods and the LLG dynamics. The core concept involves using a deep neural network to approximate $\overrightarrow{H}_{demag}$, the most computationally demanding component of the effective field, and utilizing this approximation to conduct the LLG simulation. As illustrated in Fig.~\ref{fig:frame}(c), NeuralMAG utilizes a Unet architecture comprising an encoder and a decoder, with corresponding layers of the encoder and decoder interconnected by skip connections\cite{ronneberger2015u,baccouche2021connected}. The encoder consists of a series of convolution layers that aggregate the input magnetization vectors into a hierarchy of granularity levels, where each layer's convolution kernels encode the local interaction rules at the respective granularity level. The local interactions at various levels are subsequently integrated by the decoder, leveraging the skip connections and the upsampling capabilities of the deconvolution layers. The output of the Unet, an estimation of $\overrightarrow{H}_{demag}$, serves to conduct subsequent LLG iterations, establishing a Unet/LLG iterative computation scheme.

Intuitively, the Unet decomposes the large-scale convolution in Eq.~\ref{eq:demag} into a hierarchy of small-scale convolutions, yielding an approximation for $\overrightarrow{H}_{demag}$. We demonstrate that this approximation, despite the inevitable accuracy loss, results in an algorithm with a complexity of $O(N)$ under specific conditions, significantly outperforming the FFT approach whose complexity is $O(N\log(N))$. This offers a valuable opportunity to balance efficiency and accuracy, enabling simulations for large-size models. Note that the divide-and-accumulate idea is shared by the fast multipole method (FMM) \cite{darve2000fast}, though the implementation of division and accumulation differs: it is learned by the Unet in our method, but is designed by humans in most FMM approaches.

The Unet design of NeuralMAG and the core idea behind are illustrated in Fig.\ref{fig:frame}(c). This innovative Unet/LLG iterative scheme exhibits extensive flexibility, enabling the execution of various micromagnetic simulation tasks, including the prediction of magnetic ground states and hysteresis loop forecasting. Furthermore, the framework demonstrates remarkable generalizability, which allows for training on limited small-scale data while effectively predicting the behaviour with large-scale samples, and training under randomly settled configurations while predicting for unexplored configurations, such as the sample's shape, saturation magnetization $M_s$, exchange stiffness $A_x$, and the uniaxial anisotropy energy density $K_u$. Finally, the framework can potentially leverage the latest model compression and optimization acceleration techniques offered by contemporary deep learning platforms, presenting further opportunities to continuously improve computational efficiency. The code has been packaged as a tool and is available on GitHub\footnote{See the GitHub repository at \url{https://github.com/Caiyq2019/NeuralMAG/tree/main/}}.

\section{Results}\label{sec3}

The Unet model of NeuralMAG was trained using 140k fully random samples for each of the three sizes: 32, 64, and 94. Two-thirds of these samples were 'masked' by random shapes at arbitrary locations, and the masked areas were set to zero. A random external magnetic field was also applied to these 'masked' samples. Utilizing these random samples, conventional FFT/LLG simulations were performed, and the pairs ($\overrightarrow{m}, \overrightarrow{H}_{demag}$) from every step of the simulation were collected as training data for the Unet model. After training, the Unet was employed to replace the FFT to conduct Unet/LLG simulations. We emphasize that this model is generalizable, meaning that the single model can be applied across all micromagnetic tasks based on LLG dynamics, and suitable for samples of varying sizes, shapes, and magnetic materials.

We evaluated the Unet/LLG simulation on two tasks: (1) Basic dynamics simulation to assess the accuracy of LLG iterations using the Unet approximation; (2) MH curve prediction employing LLG to determine the relaxed magnetization under a sequentially changing external magnetic field. Finally, the computation speed was analyzed to gauge the efficiency improvements.

\subsection{Basic dynamics}
\label{sec:result:basic}

The first experiment assesses the Unet/LLG approach's accuracy in simulating basic magnetization dynamics, by comparing the trajectories and ground states derived from the Unet/LLG iterations against those from conventional FFT/LLG iterations. The experiment utilizes a soft magnetic thin film as the material system, where magnetic topologies, vortex and anti-vortex \cite{shinjo2000magnetic,shigeto2002magnetic} emerge typically from a random magnetization configuration. The dynamic process primarily involves the repulsion between vortices of the same type and the attraction and subsequent annihilation between vortex/anti-vortex pairs \cite{PhysRevLett.97.177202}. Given the absence of the magnetocrystalline energy and the external field ($K_u=0$ and $H_{ext}=0$), the dynamics are governed by a competition between $H_{demag}$ and $H_{exch}$, making them particularly sensitive to the accuracy of $H_{demag}$. A set of benchmarks was established to evaluate the prediction accuracy of the ground states. These benchmarks focus on two critical characteristics of the magnetic thin film's ground state: (1) the number of vortices and (2) the orientation and polarization of these vortices. The specific methodologies employed are detailed in the Methods section.

\begin{figure}[htb!]
    \centering
    \includegraphics[width=1.0\textwidth]{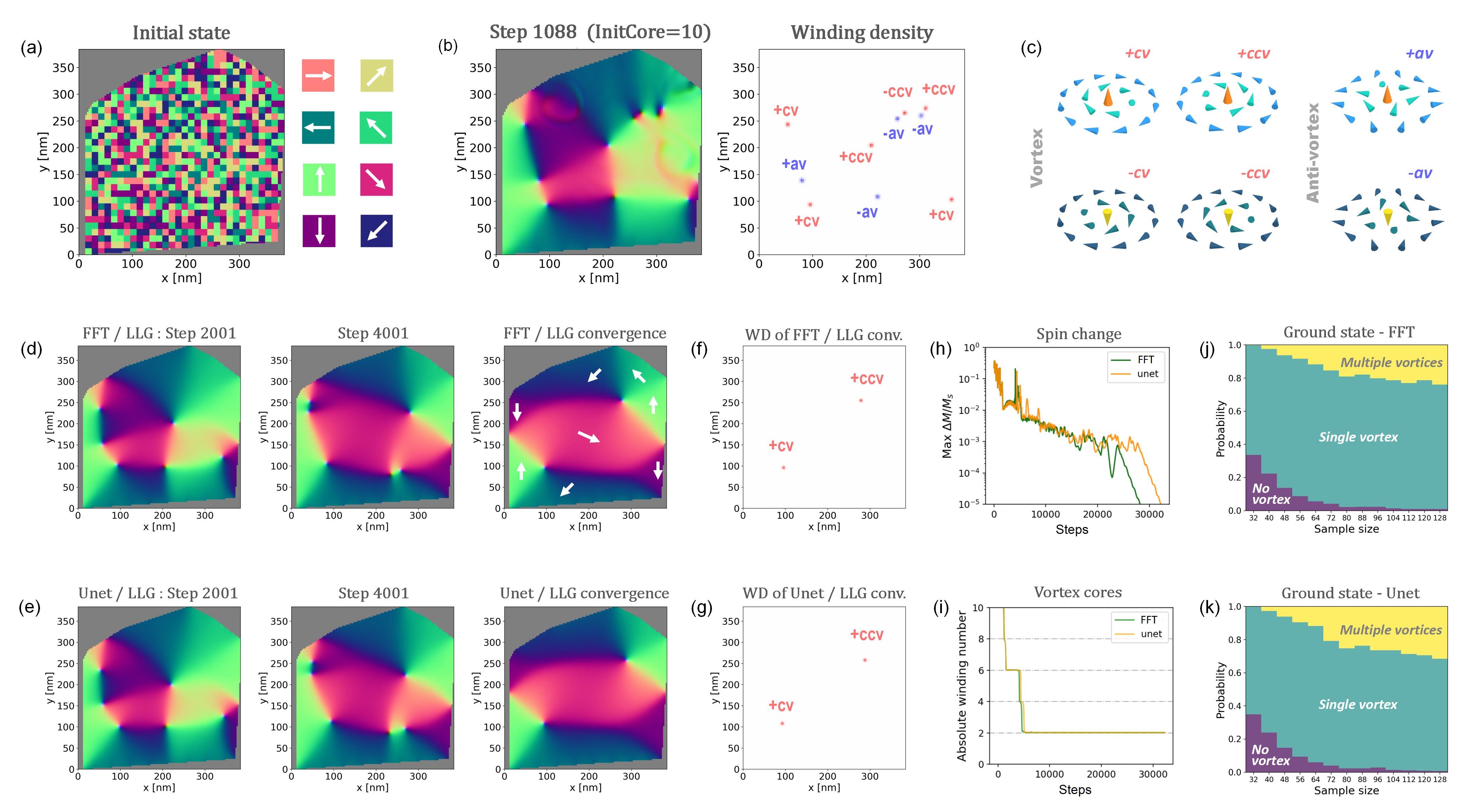}  
    \caption{
    (a)-(i) showcases basic dynamic simulations with FFT/LLG and Unet/LLG. A sample, sized at 128 and randomly shaped into a convex hull, features an FDM cell size of 3 nm, resulting in a geometric size of $384$ nm. 
    (a) The initial random state with magnetization directions indicated by different colors. 
    (b) The cooling process that converts a random initial state to a cooled state through 1088 FFT-based iterations, leading to more regular patterns and a specified number of vortices ($InitCore$), with vortex distribution depicted in the winding density plot and each vortex type labeled. (c) The definition of vortex types. (d)(e) The simulation outputs at various stages using FFT/LLG and Unet/LLG approaches, respectively. (f)(g)  Comparison on winding densities in the converged state between FFT/LLG and Unet/LLG methods. (h) Maximal spin change through LLG iterations. (i) Absolute winding numbers through LLG iterations. (j)-(k) provide phase diagrams for the ground state in square-shaped samples with default material parameters but different sizes, demonstrating the tendency of vortex cores to coexist in larger sample sizes, as shown by both FFT/LLG and Unet/LLG simulations.
    }
    \label{fig:vortex_sta}
\end{figure}

Fig.~\ref{fig:vortex_sta} {showcases} an example of the simulation with both the sample's shape and initial state randomized. We found that Unet-based simulation on fully randomized initial states often results in large prediction errors. A cooling process has been designed to solve the problem, which runs the FFT-based simulation until a relatively stable state is achieved, as {illustrated} in Fig.~\ref{fig:vortex_sta}(a)-(c). After the cooling process, the Unet can then replace FFT to {continue with} the simulation. Fig.~\ref{fig:vortex_sta}(d)-(i) {demonstrate} how the Unet-based simulation closely aligns with the outcomes of the FFT-based process.

For {quantitative analysis}, three test groups were {constructed}, each {consisting} of 100 random samples: (1) {The} Square group, {comprising} square samples with sizes ranging from 32 to 128. Importantly, the size of 128 was {not included} in the model's training set, {acting} as a test for cross-scale generalizability. (2) {The} Random Shape group, {comprising} samples shaped as irregular polygons with 30 edges, where both the size and location of these shapes are {randomized}. This group was {specifically designed} to evaluate cross-shape generalizability. (3) {The} Random Material group, where samples are uniformly square but {exhibit} variability in magnetic parameters, with saturation magnetization $M_s$ ranging from 400 to 1200 emu/cc and exchange stiffness $A_x$ {between} $3 \times 10^{-7}$ and $7 \times 10^{-7}$ erg/cm. It should be noted that the Unet model underwent training with default material settings of $M_s=1000$ emu/cc and $A_x=5 \times 10^{-7}$ erg/cm. {Consequently}, this group aims to assess cross-material generalizability.

\begin{figure}[htb!]
    \centering
    \includegraphics[width=1.0\textwidth]{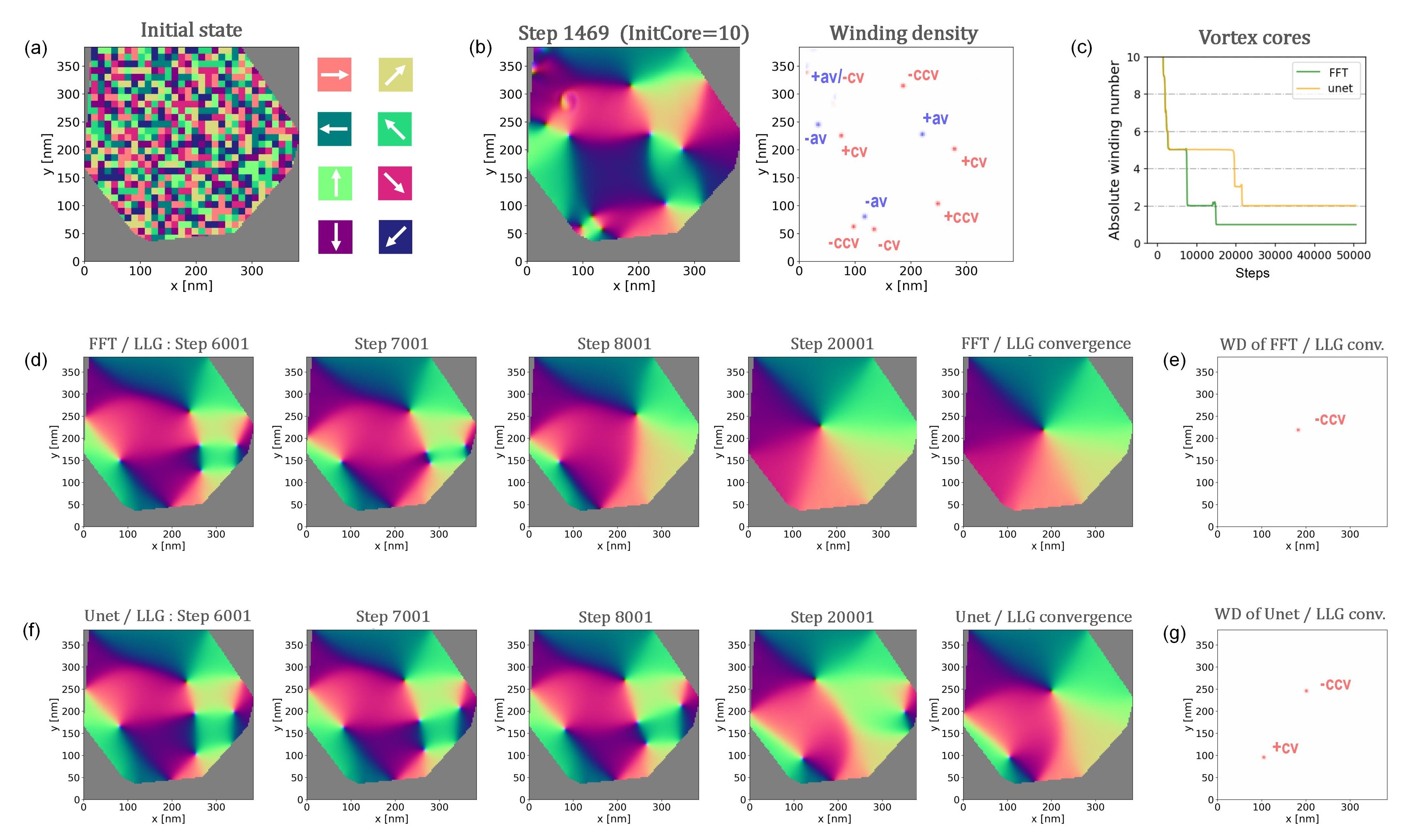}
    \caption{
    This example highlights a discrepancy in vortex number predictions by the Unet/LLG simulation. (a) The initial random state, with magnetization directions differentiated by a spectrum of colors. (b) The cooling process which transforms the random initial state into a significantly more stable configuration after 1469 FFT/LLG iterations. The vortex distribution is showcased through a winding density plot, with annotations for each vortex type. (c) Absolute winding numbers throughout the iterations between FFT/LLG and Unet/LLG simulations. (d)(f) The simulation outputs at various stages for the FFT/LLG and Unet/LLG methods, respectively. The Unet model appears to undervalue the strength of interactions between vortices, leading to delayed annihilation of vortex/anti-vortex pairs and preventing a vortex from exiting the film's edge. (e)(g) The winding density in the converged states from FFT/LLG and Unet/LLG simulations, respectively. In the FFT/LLG simulation’s converged state, only a single vortex is present, whereas two vortices are observed in the Unet/LLG simulation's final state.
    }
    \label{fig:vortex_bad1}
\end{figure}

\begin{figure}[htb!]
    \centering
    \includegraphics[width=1.0\textwidth]{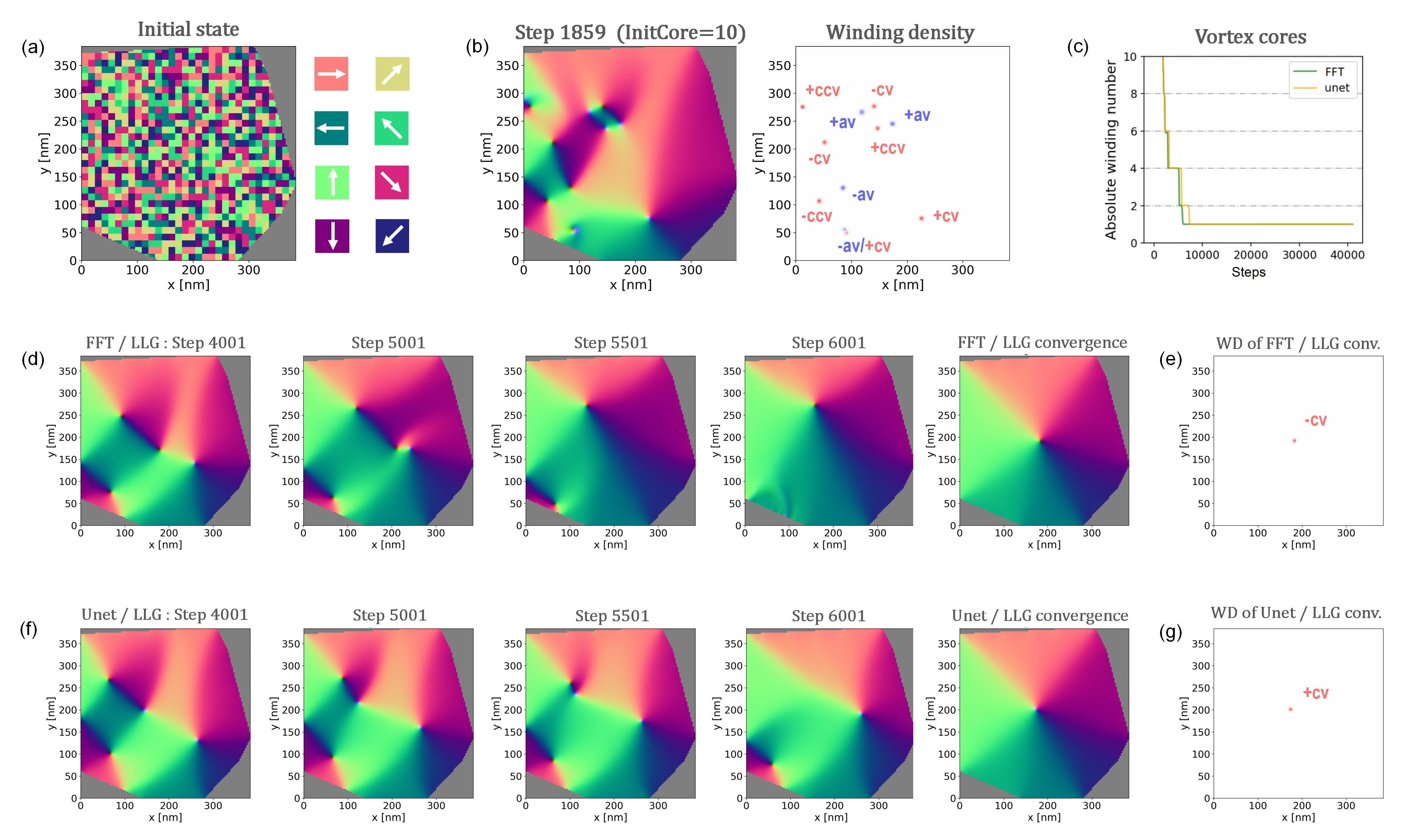}
    \caption{
    This example illustrates a specific misprediction regarding vortex properties in the Unet/LLG simulation. (a) The initial random state with magnetization directions indicated by various colors. (b) The cooling process which transforms the initial random state into a notably more stable configuration through 1859 FFT/LLG iterations. Vortex distribution is depicted via a winding density plot, with labels identifying the vortex types. (c) Absolute winding numbers between the FFT/LLG and Unet/LLG simulations throughout the iterations. (d)(f) The simulation outputs at various intervals, contrasting the FFT/LLG method with the Unet/LLG approach, respectively. Notably, at step 4001, minor discrepancies in vortex locations between FFT/LLG and Unet/LLG lead to divergent choices in vortex/anti-vortex pairs for annihilation. (e)(g) The winding density in the converged state for FFT/LLG and Unet/LLG simulations, respectively. In the FFT/LLG simulation's converged state, a negative clockwise vortex persists, in contrast to a positive clockwise vortex remaining in the Unet/LLG simulation's final state.
    }
    \label{fig:vortex_bad2}
\end{figure}

Table~\ref{tab:vortex} presents the results. It shows that with the Unet model, most prediction accuracies {surpassed} 90\%, underscoring the model's effectiveness. Specifically, the model demonstrated remarkable cross-shape and cross-material capabilities, with almost no loss in accuracy observed in the tests with varied shapes and unknown material parameters. A minor decrease in prediction accuracy was noted in the cross-scale tests (sample size 128), particularly within the random shape group, where precision fell to 89\% for vortex number and to 86\% for vortex properties, respectively. Examples of mispredictions can be {found} in Fig.\ref{fig:vortex_bad1} and Fig.\ref{fig:vortex_bad2} within the Extended Data. These observations indicate that one should cautiously interpret a single simulation result from Unet/LLG due to potential errors, although the likelihood of such errors is relatively low. However, for tasks necessitating statistical analysis of numerous simulations, such as generating a phase diagram of convergent states illustrated in Fig.~\ref{fig:vortex_sta}(j)(k), Unet/LLG can be used as a reliable tool. In this scenario, the impact of individual errors is attenuated, and the advantage of Unet/LLG in speed and memory (as detailed in Sec \ref{sec:result:speed}) becomes prominent.

\begin{table}[h]
\centering
\caption{
The prediction accuracy of the Unet/LLG simulation for the ground state of a soft magnetic film, with the cooling process initialized at $InitCore=5$.}
\label{tab:vortex}
\begin{tabular}{@{}cm{2cm}m{2cm}m{2cm}@{}}
\toprule
\textbf{Group} & \textbf{Sample Size} & \textbf{\begin{tabular}[c]{@{}c@{}} vortex number \\ precision\end{tabular}} & \textbf{\begin{tabular}[c]{@{}c@{}} vortex property \\ precision\end{tabular}} \\
\midrule
\multirow{3}{*}{Square}  & 32  & 97.00\% & 94.00\% \\
                         & 64  & 97.00\% & 97.00\% \\
                         & 128 & 97.00\% & 93.00\% \\
\hline
\multirow{3}{*}{Random Shape}  & 32  & 98.00\% & 93.00\% \\
                               & 64  & 99.00\% & 97.00\% \\
                               & 128 & 89.00\% & 86.00\% \\
\hline  
\multirow{3}{*}{Random material}  & 32 & 98.00\% & 98.00\% \\
                                 & 64 & 98.00\% & 98.00\% \\
                                 & 128 & 92.00\% & 91.00\% \\

\bottomrule
\end{tabular}
\end{table}

\subsection{MH curve}
\label{sec:result:mh}

The second experiment {involves simulating} the magnetization-field (MH) curve of a magnetic thin film. Contrary to the first experiment which began with randomized initial conditions, this experiment starts from a saturated state, meaning all magnetization vectors are aligned in one direction due to a strong external magnetic field ($H_{ext}$). Subsequently, the external field is sequentially weakened, then reversed in direction and increased again, at each step with the state stabilized before the field is changed. The average magnetization along the field direction is plotted {against} the field strength, illustrating the magnetic material's response to changes in its environment. MH curves hold significant importance in both practical applications and theoretical analysis. In practical applications, experimental measurements and numerical simulations of MH curves are crucial for designing magnetic recording media and rare-earth permanent magnets\cite{weller2016fept,nakamura2018current}. Theoretically, the MH curve gives insights into the energy landscape of the magnetic system under varying external fields ($H_{ext}$), showcasing how magnetization configurations adapt to remain at the energy minimum\cite{stoner1948mechanism}. Thus, distinct from the first experiment's emphasis on LLG dynamics, this MH curve experiment assesses the accuracy of the Unet model in predicting the energy contributions of magnetostatic interactions.

\begin{figure}[hb!]
    \centering
    \includegraphics[width=1.0\textwidth]{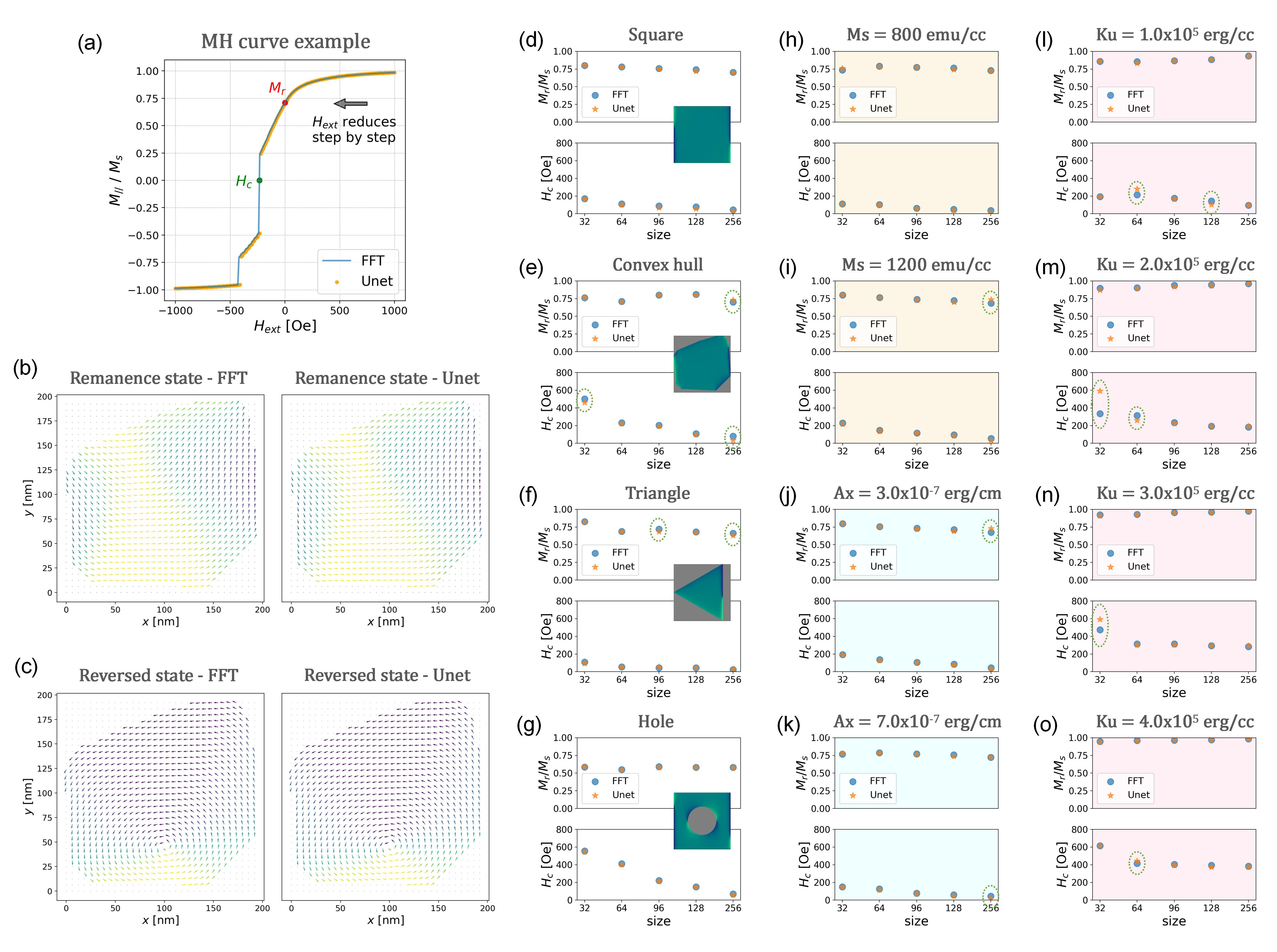}
    \caption{ 
    (a) An example of MH prediction, where the sample is modeled as a convex hull, randomly generated with a size of 64 and utilizing default material parameters. Notably, predictions from the Unet/LLG simulation align precisely with results from the conventional FFT/LLG method. 
    (b) The remanence states resulting from the MH curves in (a), as determined by both FFT/LLG and Unet/LLG simulations.
    (c) Reversed magnetic states from the same test in (a), as observed by FFT/LLG and Unet/LLG simulations immediately after the external field $H_{ext}$ surpasses the coercivity $H_c$.
    (d)-(g) Variations in the remanence ($M_r$) and coercivity ($H_c$) values across different sample shapes.
    (h)-(i) The effects of the material's saturation magnetization ($M_s$) on the $M_r$ and $H_c$ values.
    (j)-(m) The effects of the uniaxial anisotropy energy density ($K_u$) on the $M_r$ and $H_c$ values.
    (n)-(o) The effects of the exchange stiffness ($A_x$) on the $M_r$ and $H_c$ values. 
    Data points exhibiting significant discrepancies, defined by a coercivity difference $\Delta H_c \geq 25$ Oe or a relative remanence change $\Delta M_r/M_s \geq 0.03$, are distinctly highlighted with circles in figures (d)-(o).}
    \label{fig:MH}
\end{figure}

An example of the MH prediction is {presented} in Fig.\ref{fig:MH}(a)-(c). Generally, two aspects of the MH curves are mostly interesting to researchers: (1) a smooth transition from saturation to $H_{ext}=0$, and (2) a sharp magnetization reversal within a narrow external field range (magnetization reversal). Two critical points on the curve, $M_r$ (remanence) and $H_c$ (coercivity), are usually adopted to characterize these two aspects, respectively\cite{schrefl1994remanence}. $M_r$, or remanence, is the magnetization at $H_{ext}=0$, and $H_c$, or coercivity, is the external field required to reduce the magnetization to zero. These two points are used to quantify the prediction accuracy of the Unet/LLG approach for MH tasks relative to the FFT/LLG approach. As in the previous section, this experiment also emphasizes the cross-scale, cross-shape, and cross-material generalizability.

The outcomes of the MH curve experiment are depicted in Fig.\ref{fig:MH}(d)-(o), with additional details provided in Fig.\ref{fig:MHall} within the Extended Data. Of the 60 simulation cases, 49 showed a strong agreement between Unet/LLG and FFT/LLG, with no significant discrepancies in either coercivity or remanence (defined as $\Delta H_c \geq 25$ Oe or $\Delta M_r \geq 0.03$), resulting in an overall accuracy of 82\%. Focusing solely on one of the features, the prediction accuracies are 92\% for remanence and 85\% for coercivity. These findings confirm the Unet/LLG method's capacity in MH tasks, in particular its generalizability as the test involves complex settings 
on sample size, shape and material.

A notable observation is the improved generalizability with a large magnetocrystalline energy $K_u$. Severe mismatch exists when $K_u$ is small, in particular with small sizes. When $K_u$ increases to $3.0 \times 10^5$ erg/cc, severe mismatch is suppressed, remaining only in the size of 32, and it is further suppressed in all sample sizes when $K_u$ increases to $4.0 \times 10^5$ erg/cc. The behavior can be attributed to the increasing contribution of the size-invariant $H_{aniso}$ when $K_u$ is large. This characteristic underscores the feasibility of our Unet/LLG approach for MH tasks involving larger sizes and nonzero $K_u$ values, which holds practical significance for the study of full-size magnetic devices, such as reading sensors or storage components.

\begin{figure}[hb!]
    \centering
    \includegraphics[width=0.9\textwidth]{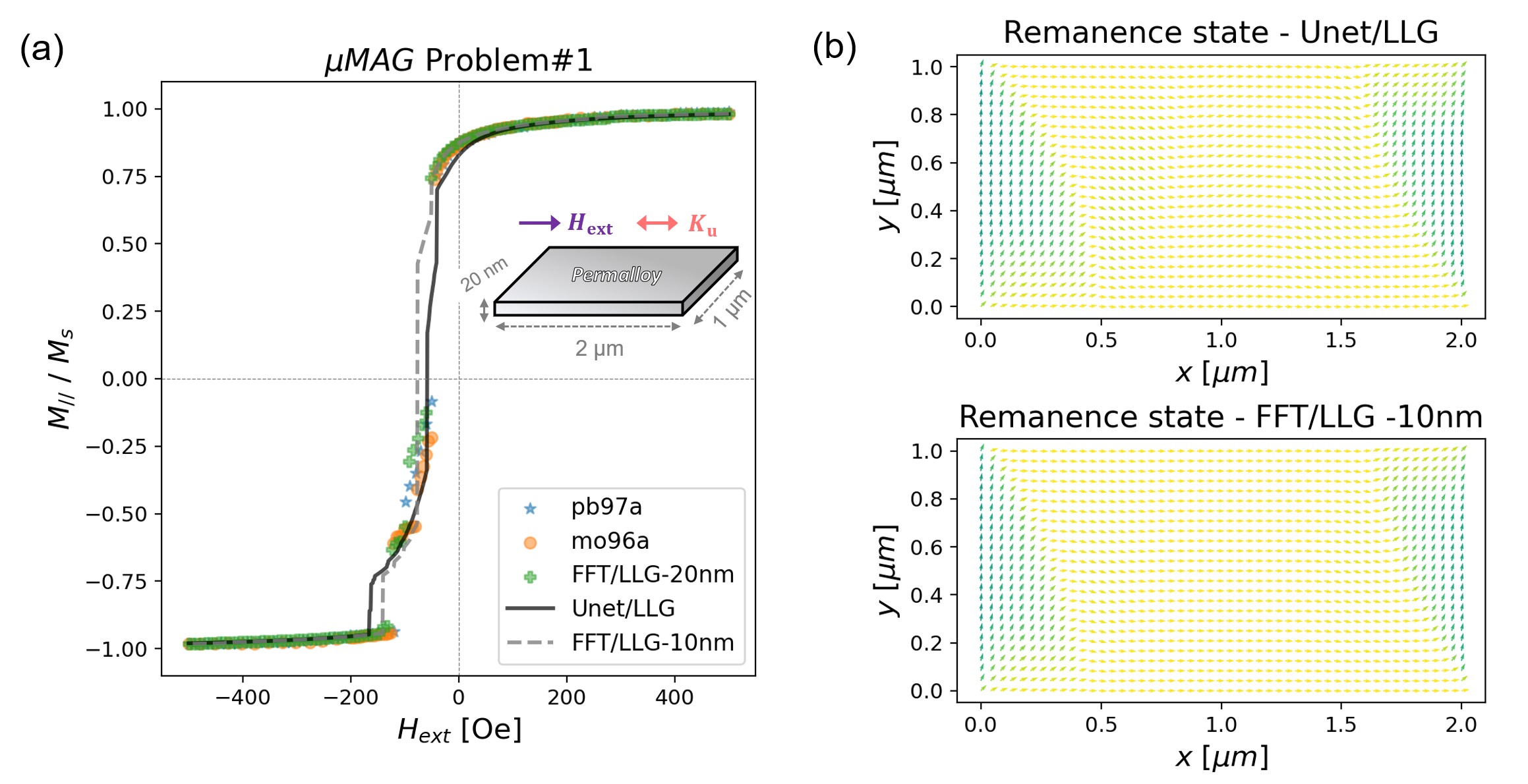}
    \caption{
    A simulation test on the $\mu MAG$ standard problem $\#1$. (a) The MH curve, simulated using the Unet/LLG approach, is depicted with a black line. For comparison, the reported data for problem $\#1$, as detailed on the $\mu MAG$ website, are displayed as scattered marks. An insert picture provides a succinct overview of the standard problem $\#1$. For the Unet/LLG simulation, an FDM model of size $256 \times 256 \times 2$ is constructed, selectively masked to preserve a rectangular region of $200 \times 100 \times 2$. Each cell within the model measures $10\text{nm} \times 10\text{nm} \times 10\text{nm}$. The assumed magnetic parameters for the permalloy include a saturation magnetization ($M_s$) of $800$ emu/cc, exchange stiffness ($A_x$) of $1.3 \times 10^{-6}$ erg/cm, and uniaxial anisotropy energy density ($K_u$) of $5000$ erg/cc. (b) The magnetization configuration for problem \#1 at the remanence state. The upper configuration was generated by Unet/LLG and the lower one is a reproduction of the "mo96a"-series data reported on the $\mu MAG$ website.
    }
    \label{fig:prb1}
\end{figure}

To showcase the Unet/LLG method's broad applicability to MH tasks, we examine the performance of the Unet model on the $\mu$MAG standard problem \#1, focusing on the MH curve of a rectangular permalloy film, as depicted in Fig. \ref{fig:prb1}. Significantly, the simulation settings, encompassing sample size, shape, magnetic properties, and unit cell size, are all different from those used in the training set. Thus, this represents a thorough examination of the model's cross-scale, cross-shape, and cross-material generalizability. As illustrated in Fig. \ref{fig:prb1}(a), the MH curve simulated by the Unet/LLG closely aligns with the benchmarks across most of the field range. Specifically, for the remanence state, the magnetization pattern produced by the Unet/LLG closely matches the reported data, as demonstrated in Fig. \ref{fig:prb1}(b).

\subsection{Computational Efficiency Evaluation}
\label{sec:result:speed}

Micromagnetic simulation, which explores the magnetic behavior of materials at the microscale, is notably resource-intensive, particularly for large-scale samples. Consequently, there is a critical need to enhance computational efficiency and reduce memory consumption. Our study compares the performance of three simulation methodologies: FFT/LLG, Unet/LLG, and Unet/LLG accelerated by TensorRT\footnote{\url{https://pytorch.org/TensorRT/}}, with results evaluated on a Nvidia RTX-3090 GPU.

Table \ref{tab:compute_consuming} presents a comparative analysis of the conventional FFT/LLG approach versus the TensorRT-accelerated Unet/LLG model, focusing on computation speed and GPU memory usage. With increasing sample size, the TensorRT-accelerated Unet/LLG approach is significantly faster than the FFT/LLG method. Remarkably, for a sample size of 2048, the Unet/LLG method realized a 6.7-fold enhancement in computation speed, alongside a reduction to 0.53 times its original memory consumption.

The improved speed and decreased memory requirements of the TensorRT-accelerated Unet/LLG model compared to the FFT/LLG approach stem the lower computational complexity $O(N)$, as well as the advanced optimization capabilities of the TensorRT framework integrated with the PyTorch ecosystem. TensorRT's optimization techniques, including mixed-precision inference, layer fusion, and kernel auto-tuning, facilitate the efficient execution of deep learning models. These methods contribute to significant enhancements in computational speed and memory efficiency.

Table \ref{tab:speed_all} in the Extended Data provides a more detailed comparison of the computational efficiency of three simulation methodologies: FFT/LLG, Unet/LLG, and Unet/LLG accelerated by TensorRT. The analysis reveals that the Unet/LLG model, even without TensorRT, offers a significant speed advantage over the conventional FFT/LLG approach, especially when the sample size is large. This advantage demonstrates the benefit of the lower computational complexity $O(N)$ with UNet compared to $O(N\log(N)$ with FFT. It is further magnified by incorporating TensorRT acceleration. This is particularly evident in the marked reduction in the $H_{demag}$ computation time per iteration, showcasing the considerable efficiency gains achieved through UNet and TensorRT acceleration.
The advancements on speed and resource consumption enable simulations of large samples.

\begin{table}[htb!]
\centering
\caption{
Analyzing computational speed and GPU memory usage: a comparative study of FFT/LLG versus TensorRT-accelerated Unet/LLG on an Nvidia RTX-3090 GPU.}
\label{tab:compute_consuming}
\begin{tabular}{@{}c c c c c c c c c@{}}
\toprule
\multirow{2}{*}{\textbf{size}} &
  \multicolumn{3}{c}{\textbf{FFT}} &
  \multicolumn{3}{c}{\textbf{Unet-TensorRT}} &
  \multirow{2}{*}{\textbf{\begin{tabular}[c]{@{}c@{}}Speed \\ ratio\end{tabular}}} &
  \multirow{2}{*}{\textbf{\begin{tabular}[c]{@{}c@{}}Mem \\ ratio\end{tabular}}} \\
\cmidrule(lr){2-4} \cmidrule(lr){5-7}
 &
  \textbf{\begin{tabular}[c]{@{}c@{}}Iteration \\ time\end{tabular}} &
  \textbf{GPU/MiB} &
  \textbf{\begin{tabular}[c]{@{}c@{}}$H_{demag}$ time\\ per iteration\\\end{tabular}} &
  \textbf{\begin{tabular}[c]{@{}c@{}}Iteration \\ time\end{tabular}} &
  \textbf{GPU/MiB} &
  \textbf{\begin{tabular}[c]{@{}c@{}}$H_{demag}$ time\\ per iteration\end{tabular}} &
   &
   \\
\midrule
32   & 5.80E-03 s & 344 & 2.28E-03 s & 5.40E-03 s & 354   & 1.56E-03 s & 1.5 & 1.03 \\
128  & 5.90E-03 s & 386 & 2.36E-03 s & 5.50E-03 s & 380   & 1.60E-03 s & 1.5 & 0.98 \\
512  & 2.00E-02 s & 1062 & 1.32E-02 s & 1.00E-02 s & 732   & 3.40E-03 s & 3.9 & 0.69 \\
1024 & 8.10E-02 s & 3286 & 5.60E-02 s & 3.70E-02 s & 1914  & 1.04E-02 s & 5.4 & 0.58 \\
2048 & 3.60E-01 s & 12130 & 2.56E-01 s & 1.50E-01 s & 6474  & 3.80E-02 s & 6.7 & 0.53 \\
3072 & -- & -- & -- & 3.30E-01 s & 14122 & 8.40E-02 s & -- & -- \\
\bottomrule
\end{tabular}
\end{table}

\section{Discussion}
\label{sec:disc}

\subsection{Generalizability}
\label{sec:disc:gen}

A key strength of the Unet/LLG approach compared to existing DL-based methods is its exceptional generalizability, i.e., a single model is capable of handling diverse tasks across samples of varying sizes, shapes, and material properties, establishing it as a versatile tool for micromagnetic simulation. Theoretically, this generalizability originates from the universal physical rule governing the cross-cell interaction during demagnetization. More specifically, regardless of the cell's location, the cross-cell interaction follows the same principle and so can be calculated with the same procedure, conditioned on a set of parameters determined by the material's physical characteristics.

Technically, {this capability is derived from} two {pivotal} designs within NeuralMAG: the Unet/LLG nested framework and the {implementation of local learning in Unet}. In the Unet/LLG framework, the Unet component is {specifically responsible for modeling} the cross-cell interactions, while the {evolutionary} dynamics and material-specific settings are {managed by the LLG iterations}. This arrangement {implies that if the physical rule governing cross-cell interaction is universally applicable, then Unet/LLG framework possesses inherent generalizability concerning material settings and can be applied to any task that relies on LLG dynamics, i.e., task and material generalizability.}

The ability to generalize across sample sizes and shapes is rooted in its local learning feature of the Unet model. Let's delve into the Unet architecture to understand how it computes the demagnetizing field. In the first convolution layer, the interaction among primary cells is represented by the convolution kernels, with a limited kernel size of 3 capturing only local interactions. To address long-range interactions, the encoder progressively increases the receptive field of the neurons through downsampling, enabling the learning of local interactions on a broader scale, as depicted in Fig.~\ref{fig:frame}(c). Within each hidden layer, 2 x 2 neighboring cells are consolidated into a single cell, establishing a new granularity level that differs in cross-cell interactions from previous levels. The convolution kernels, operating at this enlarged granularity, are expected to capture larger-scale interactions while still maintaining local learning due to the limited kernel size.
The decoder aggregates local interactions across various granularities by extensively utilizing skip connections. Our fundamental assumption is that by aggregating local interactions at different levels of granularity, the Unet model can approximate global cross-cell interactions. The accuracy of this approximation will be discussed later; for now, our focus is on elucidating how the Unet architecture enables size and shape generalizability.

Firstly, it's important to note that the local interaction rule is independent of shape, making the Unet model inherently capable of generalizing across different shapes. Additionally, since samples of any size can be broken down into small blocks, and all these blocks adhere to the same physical rules for local interaction, the entire demagnetization field can be calculated on a block-by-block basis, ensuring generalizability across different sizes. This block-by-block processing is illustrated in the fourth hidden layer of the Unet architecture, as depicted in Fig.~\ref{fig:frame}(c), where the size of the feature maps scales linearly with the size of the input samples.

\subsection{Bound of accuracy}
\label{sec:disc:acc}

As {previously mentioned}, Unet is {designed} to approximate global inter-cell interactions in the {computation} of the demagnetizing field by {aggregating} local interactions across multiple granularity levels, from coarse to fine. {Firstly notice that} deep CNN models are universal approximators\cite{zhou2020universality}, {which means} that with a sufficiently complex Unet structure, the accuracy of approximation is {assured}, at least for sample sizes represented in the training dataset. However, to {maintain} efficiency, the model structure needs to be compact, for instance the Unet architecture depicted in Fig.~\ref{fig:frame}(c). We will discuss the upper bound of accuracy achievable with this design.

The first scenario we will discuss occurs when the size of the input sample does not exceed the receptive field of a single neuron in the bottleneck layer, specified as 16 in Fig.~\ref{fig:frame}(c). Under these conditions, the dependencies between any two cells within the sample can be accurately represented, leading to effectively modeling the global inter-cell interaction. Therefore, prediction accuracy is assured, provided that the convolutional kernels are adequately comprehensive and the learning process perfectly converges.

The second scenario arises when the sample size is larger than the receptive field of neurons in the bottleneck layer. Under these conditions, the feature maps in the bottleneck layer do not consist of singular points but rather contain multiple neurons. The interactions among these bottleneck neurons are inadequately represented (even though some neighboring neurons may interact via the deconvolution kernel), indicating a lack of capacity to accurately depict the interactions between \emph{any} two cells in the sample. Therefore, Unet serves as an \emph{imperfect} approximator in this scenario, with the approximation deteriorating as the sample size increases.

To enhance accuracy for large samples, one possible approach is to increase the depth of the encoder. This approach is feasible as the computational complexity is $O(N)$ if the receptive field of the pooling operations is sufficiently large, a point that will be elaborated on shortly. However, this approach requires the generation of corresponding training samples via traditional FFT-based simulation, a process currently not feasible due to its high cost. A potential solution could involve developing a dynamic network that tailors the Unet structure based on sample size, utilizing shared convolution kernels derived from small-scale samples. However, this approach might result in further accuracy degradation. We leave this line of investigation for future research.

\subsection{Complexity}
\label{sec:disc:complex}

We can begin by analyzing the computational complexity of the encoder in the Unet model. Let's assume that all the convolution kernels are of size $r$ with a stride of 1. Additionally, suppose the size of the input samples is $N=n \times n$, and the first hidden layer has $c$ channels. As we progress through each subsequent convolution layer, the resolution decreases by a factor of $1/\alpha$, while the number of channels increases by a factor of $\beta$. This sequential reduction in feature map size leads to a single 1 x 1 map over $\log_{\alpha^2} N$ steps. The number of multiplication operations involved in this encoding process can be calculated as follows:

\begin{eqnarray}
\begin{aligned}
&6r^2N \times c &  [Input] \\
+& r^2  c \frac{N}{\alpha^2} \times \beta c & [H1]  \\
+ &r^2 \beta c \frac{N}{\alpha^4} \times \beta^2 c & [H2]   \\
+&... & \\
+ &r^2 c^2 N \frac{\beta^{2k-1}}{\alpha^{2k}} & [H_k]   \\
+&... &  \\
+&r^2 c^2 N \frac{\beta^{2 \log_{\alpha^2}N -1}}{\alpha^{2 \log_{\alpha^2}N}} & [H_{\log_{\alpha^2}N}]
\end{aligned}
\end{eqnarray}

\noindent Simple calculation shows that the computation amounts to:

\begin{equation}
6r^2N \times c + r^2c^2\frac{\beta}{\alpha^2}N\left\{1+ \left[\frac{\beta^2}{\alpha^2}\right]^1+...+\left[\frac{\beta^2}{\alpha^2}\right]^{log_{\alpha^2}N-1}\right\}
\end{equation}

In the scenario where $\alpha = \beta$, the computational complexity can straightforwardly be demonstrated as $O(N\log N)$, given that the summation includes $\log N/\log \alpha^2$ items. For cases where $\alpha \ne \beta$, the calculations within the brackets form a geometric series with a common ratio $q=\beta^2/\alpha^2$, resulting in the computation being expressed as follows:

\begin{equation}
6r^2N \times c + r^2c^2\frac{\beta}{\alpha^2}N \frac{1-q^{\log_{\alpha^2}N}}{1-q}
\end{equation}

\noindent If $\beta<\alpha$, the above computation is bounded by 
\begin{equation}
6r^2N \times c + r^2c^2\frac{\beta}{\alpha^2}N \frac{1}{1-q},
\end{equation}
which means the complexity is $O(N)$.
If $\beta > \alpha$, the computation is:

\begin{equation}
\begin{aligned}
&6r^2N \times c + r^2c^2\frac{\beta}{\alpha^2}N \frac{q^{\log_{\alpha^2}N} -1}{q-1} \\
<& 6r^2N \times c + r^2c^2\frac{\beta}{\alpha^2}N \frac{1}{q-1}[q^{\log_{\alpha^2}N}] \\
=&6r^2N \times c + r^2c^2\frac{\beta}{\alpha^2} \frac{1}{q-1}[(\beta^2)^{\log_{\alpha^2}N}] \\
=&6r^2N \times c + r^2c^2\frac{\beta}{\alpha^2} \frac{1}{q-1}[(\beta^2)^{\frac{\log_{\beta^2}N}{\log_{\beta^2}\alpha^2}}] \\
=&6r^2N \times c + r^2c^2\frac{\beta}{\alpha^2} \frac{1}{q-1}[N^{\frac{1}{\log_{\beta^2}\alpha^2}}] \\
=&6r^2N \times c + r^2c^2\frac{\beta}{\alpha^2} \frac{1}{q-1}[N^{\frac{\log \beta^2}{\log \alpha^2}}]
\end{aligned}
\end{equation}

\noindent Therefore, the computational complexity is $O(N^{\frac{\log \beta}{\log \alpha}})$, which is higher than $O(N)$ with $\beta > \alpha$.

Finally, the decoder's complexity is twice that of the encoder's, a consequence of concatenating the feature maps through skip connections. In total, the computational complexity of the Unet model aligns with that of the encoder.

The derivation previously discussed utilizes a pyramid architecture, which systematically reduces feature maps to singular points, creating a `complete' structure. This design guarantees interaction between each pair of cells. More commonly, a truncated architecture, as illustrated in Fig.~\ref{fig:frame}, is employed, offering a straightforward $O(N)$ complexity, irrespective of the values of $\alpha$ and $\beta$. This truncated architecture trades a degree of accuracy for increased computational speed. Additionally, considering that FFT's complexity is $O(N\log N)$, our method—whether adopting the truncated or the pyramid version with $\alpha > \beta$, theoretically surpassing the FFT-based approach in speed, particularly as $N$ increases.

\section{Methods}
\label{sec:method}

\subsection{Micromagnetic simulation}
\label{sec:method:intro}

Following the Landau-Lifshitz-Gilbert (LLG) dynamical equation, as presented in Eq.(\ref{eq:llg}), the evolutionary process of a micromagnetic system can be numerically simulated by partitioning the material into small units and calculating magnetization within each unit. This computation unfolds iteratively, yielding a detailed temporal evolution of the system. Two popular simulation methods are the finite-differential scheme\cite{victora1987quantitative} and the finite-element scheme\cite{fidler2000micromagnetic}. In this study, we adopt the finite-differential scheme.

In the finite-differential scheme, a magnetic material sample is discretized into a regular lattice, with each cell assigned a magnetic moment \( \vec{m} \), i.e., the magnetization vector \( \vec{M} \) divided by its saturation magnetization \( M_s \). Determining the state of a sample means knowing the directions of \( \vec{m}_{ijk} \) for every cell. The evolution of \( \vec{m}_{ijk} \) is primarily governed  by the local energies and their interactions, including Zeeman energy \( E_z \) as a response to the applied field \( H_{ext} \), magnetocrystalline anisotropy energy \( E_{aniso} \), Heisenberg exchange interaction \( E_{exch} \) among neighbors, and magnetostatic/demagnetizing energy \( E_{demag} \) between pairs of \( \vec{m}_{ijk} \). Following Landau's suggestion, energy variation in each cell creates an effective field \( \vec{H}_{eff} = - \delta E / \delta \vec{m}M_s \), and the temporal evolution of the magnetic moment can be described by the LLG equation, as shown in Eq.(\ref{eq:llg}) or reformed below:

\begin{equation}
\label{eq:llg2}
d \vec{m} / dt = \vec{H}_{eff} - \vec{m} (\vec{H}_{eff} \cdot \vec{m}).
\end{equation}

The effective field \( \vec{H}_{eff} \) is calculated as the sum of several components: the external field \( \vec{H}_{ext} \), the exchange field \( \vec{H}_{exch} \), the anisotropy field \( \vec{H}_{aniso} \), and the demagnetizing field \( \vec{H}_{demag} \), as presented in Eq.(\ref{eq:eff}) and relisted below:

The effective field \( \vec{H}_{eff} \) is computed by summing up the external field \( \vec{H}_{ext} \), the exchange field \( \vec{H}_{exch} \), the anisotropy field \( \vec{H}_{aniso} \), and the demagnetizing field \( \vec{H}_{demag} \), formulated by Eq.(\ref{eq:eff}) and reproduced as below:

\begin{equation}
\label{eq:Heff}
\overrightarrow{H}_{eff} = \overrightarrow{H}_{ext} + \overrightarrow{H}_{exch} + \overrightarrow{H}_{aniso} + \overrightarrow{H}_{demag} ,
\end{equation}

\noindent where 

\begin{equation}
\label{eq:Haniso}
\overrightarrow{H}_{aniso} = H_k \cdot (\vec{m}_{ijk} \cdot \vec{k}) \vec{k} \text{ \ \ for uniaxial anisotropy}
\end{equation}

\begin{equation}
\label{eq:Hexch}
\overrightarrow{H}_{exch} = H_A \cdot \sum_{lmn}(\vec{m}_{lmn} - \vec{m}_{ijk}), \ \  l=i\pm 1 \ \ m=j\pm 1 \ \  n=k\pm 1
\end{equation}

\begin{equation}
\label{eq:Hdemag}
\overrightarrow{H}_{demag}(i,j,k) = M_s \cdot \sum_{lmn} \Omega(l-i, m-j, n-k) \cdot \overrightarrow{m}_{lmn}
\end{equation}

The exchange field constant, denoted by $H_A = 2A_x / M_s D^2$, involves $A_x$ as the exchange stiffness, $M_s$ as the saturation magnetization, and $D$ as the size of the finite-differential cell. The anisotropy field constant, expressed as $H_K = 2K_u / M_s$, involves $K_u$ as the uniaxial anisotropy energy density. The demagnetizing tensor $\Omega(\cdot,\cdot,\cdot)$ represents the interaction between cells. Convolution of this tensor with the magnetic moment $\vec{m}$ yields the demagnetization field $\vec{H}_{demag}$, constituting the most computationally demanding phase in determining $\vec{H}_{eff}$. The straightforward computation approach exhibits a time complexity of $O(N^2)$. To speed up the convolution process, the fast Fourier transformation (FFT) algorithm is typically employed, reducing the computational complexity to $O(N\log(N))$.

After calculating the gradient $d\vec{m}/dt$ based on the LLG equation, we update the magnetic moment $\vec{m}$ for each cell. For the update process in our implementation, we utilize the fourth-order Runge-Kutta (RK4) method\cite{butcher1996history}. For an in-depth explanation of this workflow, please see Algorithm \ref{alg:workflow}.

\begin{algorithm}
\caption{FDM Micromagnetic Simulation Workflow}
\label{alg:workflow}

\begin{algorithmic}[1]

\State Initialize parameters $H_A, H_K, \Omega$ with \Call{Para\_Initial}{}

\Procedure{SpinUpdate}{$\vec{m}_{ijk}$}
    \For{each cell indexed by $i, j, k$}
        \State {$\vec{D}_{ijk} \gets$ \Call{GetIncrement}{$\vec{m}_{ijk}, \Delta t$} }
        \State $\vec{m}_{ijk} \gets$ normalize($\vec{m}_{ijk} + \vec{D}_{ijk}$) 
        \Comment{Update the magnetic moment}
    \EndFor
    \State \textbf{return} $\vec{m}_{ijk}$
\EndProcedure

\Function{GetIncrement}{$\vec{m}_{ijk}, \Delta t$}
    \State $k_1 \gets$ \Call{GetGradient}{$\vec{m}_{ijk}$}
    \State $k_2 \gets$ \Call{GetGradient}{$\vec{m}_{ijk} + \frac{1}{2}\Delta t \cdot k_1$}
    \State $k_3 \gets$ \Call{GetGradient}{$\vec{m}_{ijk} + \frac{1}{2}\Delta t \cdot k_2$}
    \State $k_4 \gets$ \Call{GetGradient}{$\vec{m}_{ijk} + \Delta t \cdot k_3$}
    \State $\vec{D}_{ijk} \gets \Delta t \cdot (\frac{1}{6}k_1 + \frac{1}{3}k_2 + \frac{1}{3}k_3 + \frac{1}{6}k_4)$
    \Comment{Fourth-order Runge-Kutta}
    \State \textbf{return} $\vec{D}_{ijk}$
\EndFunction

\Function{GetGradient}{$\vec{m}_{ijk}$}
    \State $\vec{H}_{\text{ext}} \gets$ \Call{ExternalField}{$i, j, k$}
    \State $\vec{H}_{\text{exch}} \gets$ \Call{ExchangeField}{$i, j, k$}
    \State $\vec{H}_{\text{aniso}} \gets$ \Call{AnisotropyField}{$i, j, k$}
    \State $\vec{H}_{\text{demag}} \gets$ \Call{DemagField}{$i, j, k$}
    \State $\vec{H}_{\text{eff}} \gets \vec{H}_{\text{ext}} + \vec{H}_{\text{exch}} + \vec{H}_{\text{aniso}} + \vec{H}_{\text{demag}}$
    \State $\vec{G}_{ijk} \gets \vec{H}_{\text{eff}} - \vec{m}_{ijk}(\vec{H}_{\text{eff}} \cdot \vec{m}_{ijk})$ 
    \Comment{LLG equation}
    \State \textbf{return} $\vec{G}_{ijk}$
\EndFunction

\Function{ExchangeField}{$i, j, k$}
    \State \textbf{return} $H_A \cdot \sum_{lmn} (\vec{m}_{lmn} - \vec{m}_{ijk}), \quad l = i \pm 1, m = j \pm 1, n = k \pm 1$
\EndFunction

\Function{AnisotropyField}{$i, j, k$}
    \State \textbf{return} $H_K \cdot (\vec{m}_{ijk} \cdot \vec{k})\vec{k}$
\EndFunction

\Function{DemagField}{$i, j, k$}
    \State \textbf{return} $M_s \cdot \sum_{lmn} \Omega(l - i, m - j, n - k) \cdot \vec{m}_{lmn}$ \Comment{Apply FFT for efficiency}
\EndFunction
\end{algorithmic}
\end{algorithm}

\subsection{Unet Structure}
\label{sec:method:Unet}

In our research, we employed a CNN-based Unet, traditionally utilized in image segmentation\cite{ronneberger2015u}, to approximate the convolution function detailed in Eq.\ref{eq:demag}, specifically, $\vec{H}_{demag}=Unet(\vec{m})$. As depicted in Fig.~\ref{fig:frame} (c), the Unet architecture includes an encoder, consisting of a series of CNN layers designed to learn local interactions across various scales, and a decoder that aggregates these interactions to approximate the global convolution. The specifics of the model are outlined as follows.

\paragraph{Encoder module}

Unless explicitly stated otherwise, the stride for all convolution kernels is set to 1. The Unet model's encoder initiates with a 3x3 convolution layer (s0) that processes the input data that comprises a 6-channel spin $\vec{m}$ of dimensions $w \times w$, and generates the first hidden layer with 16 channels. Subsequently, the encoder undergoes a series of downsampling stages, labeled s1 to s4. Each downsampling stage comprises an average pooling operation that halves the spatial dimensions of the hidden layers, followed by a convolution block. The convolution block incorporates: (1) a LeakyReLU activation function with a negative slope of 0.2\cite{maas2013rectifier}, (2) a 3x3 convolution layer that maintains the spatial dimensions of the feature maps while doubling the channel count, (3) a layer of batch normalization\cite{ioffe2015batch}.

\paragraph{Decoder module}

The Unet model's decoder involves a series of upsampling stages (s5 to s8), each equipped with a ReLU activation\cite{nair2010rectified}, a transposed 4x4 convolutional layer with a stride of 2\cite{dumoulin2016guide}, followed by batch normalization. This 4x4 convolutional process effectively doubles the spatial dimensions of the feature map while halving the channel count. Stages s6 and s7 incorporate a dropout rate of 0.2\cite{srivastava2014dropout}, aimed at bolstering the model's capacity for generalization. The upsampling stages progressively propagate high-level interactions to finer scales at lower levels. Here, these high-level interactions are accumulated with the low-level interactions calculated at the corresponding layers in the encoder, facilitated by skip connections.

The decoder's final phase consists of two size-preserving convolution layers (s9 and s10), with layer s9 blending cell-level and high-level interactions, and layer s10 producing the 6-channel demagnetization field $\vec{H}_{demag}$. Layer s9 employs a 3x3 2D convolution, succeeded by batch normalization and a Tanh activation function. Layer s10 features a 1x1 2D convolution, reducing the channel count to 6, aligning with the target demagnetization field $\vec{H}_{demag}$. Ultimately, a specialized activation layer (s11) is devised to adjust the Unet's output to the numerical range of the demagnetizing field. This activation function is termed `inverse Symmetric Logarithmic Activation', details of which will be presented shortly.

\paragraph{Symmetric Logarithmic Activation}

To address the challenges posed by the vast value range of the demagnetizing field (\( H_d \)), spanning from \(10^{-4}\) to \(10^{4}\), we propose a novel activation function called Symmetric Logarithmic Activation (SLA) as follows:

\begin{align}
    \text{For } x \geq 0 &: \quad y = \ln(x + 1), \\
    \text{For } x < 0 &: \quad y = -\ln(-x + 1).
\end{align}

\noindent Note that SLA is an invertible function, and its inverse is given by:

\begin{align}
    \text{For } x \geq 0 &: \quad x = e^y - 1, \\
    \text{For } x < 0 &: \quad x = -e^{-y} + 1.
\end{align}

\noindent Utilizing SLA allows us to compress the demagnetizing field's values into a narrower range, thereby simplifying the Unet training process. In contrast, employing the inverse Symmetric Logarithmic Activation (ISLA) enables us to expand the Unet's output back to the actual scale of the demagnetizing field necessary for LLG simulation.

\paragraph{Weight Initialization}

All layers within the Unet architecture are initialized utilizing the Kaiming normalization technique\cite{he2015delving}. The convolutional layers adopt the `fan\_out' initialization mode, whereas the transposed convolutional layers are initialized using the `fan\_in' mode. Batch normalization layers are specifically initialized with ones for weights and zeros for biases.

\subsection{Data Generation}
\label{sec:method:dt}

In our research, we model the demagnetizing field $H_{demag}$ within the magnetic thin films utilizing Unet. For training this model, it is essential to gather data from the steps of FFT/LLG simulations. The goal of the training is to enforce the Unet model to produce the same demagnetizing field as FFT under various spin states.

We consider a bilayer thin-film structure with a cell size set to \(3nm \times 3nm \times 3nm\). The default simulation settings are outlined as follows:

\begin{itemize}
    \item Saturation magnetization \(M_s = 1000\) emu/cc,
    \item Exchange constant \(A_x = 0.5 \times 10^{-6}\) erg/cm,
    \item Uniaxial anisotropy constant \(K_u = 0.0\) erg/cc,
    \item Damping coefficient \(\lambda = 0.1\).
    \item Time step, $\Delta t = 1.0 \times 10^{-13}$ s,
    \item Convergence threshold, $(\Delta m)_{max} \leq 1.0 \times 10^{-5}$.
\end{itemize}

To train the Unet model, we perform FFT-based simulation and collect the training data from the simulation trajectories. The details are as follows.

\textbf{Random Size and Spin:}
For each simulation, we model a bilayer film with an in-plane size $w$ which represents the film's width and length and can be 32, 64, or 96. The simulation starts with a random magnetization configuration whose size is $w^2 \times layers \times channels$, where $layers=2$ reflects the two layers of the film, and $channels=3$ reflects the 3 spacial directions of the magnetization. It's important to note that the magnetization's initial state is completely randomized within the film plane, in order to generate diverse and intricate data for both training and testing. For each specified size, we conducted 100 simulations, each characterized by a unique initial magnetization configuration.

\textbf{Random External Magnetic Field:} 
For each simulation, a random external magnetic field is applied, with its magnitude ranging between 100-1000 Oe and its orientation randomly distributed within the film plane. Systems with larger magnetic fields tend to converge faster. The randomization of the external magnetic field settings helps balance the dataset between non-stable and stable states.

\textbf{Random Shape Masking:} 
To enhance the complexity of the data produced, part of the samples in the simulation were masked by a shape mask matrix. With the mask matrix applied, the magnetization of the non-masked cells were set to zero, meaning that non-magnetic substance exists in the area. The masking shapes were polygons whose location and sizes were randomly set. In our experiment, about 2/3 of the simulations employed random shape masking. Training on randomly masked samples enables the Unet model to achieve cross-shape generalization.

\textbf{Simulation:} 
Starting with the initialized magnetization configuration (possibly masked) and the randomized external magnetic field, the FFT-based simulation was conducted following the LLG dynamics. The time step of each iteration of the simulation is $10^{-13}$ s, and the simulation ran iteratively until convergence, for which the convergence criterion is $(\Delta m)_{max} \leq 1.0 \times 10^{-5}$.

\textbf{Training data Collection:} 
From each simulation, 500 pairs of $(\vec{m}, \vec{H}_{demag})$  were randomly selected as training data for the Unet model. There were 300 simulations in total, amounting to 140k training and 10k validation samples for each of the three sample sizes: 32, 64 and 96.

\subsection{Training method}
\label{sec:emthod:training}

\paragraph{Training objective}
Leveraging a total of 420k training samples (140k for each size: 32, 64, and 96), the Unet model was trained with the Mean Squared Error (MSE) loss function, formally written as follows:


\begin{equation}
\mathcal{L}(\theta) = \left( \frac{1}{N} \sum_{i=1}^{N} \left( \text{ISLA}\left( f_\theta(\vec{m}_i) \right) - H_{demag_i} \right)^2 \right) + \lambda \left( \frac{1}{N} \sum_{i=1}^{N} \left( f_\theta(\vec{m}_i) - \text{SLA}\left( H_{demag_i} \right) \right)^2 \right)
\end{equation}

\noindent Where $N$ represents the total number of training samples, and $f_\theta(\cdot)$ denotes the Unet's mapping function parameterized by $\theta$. The function $SLA(\cdot)$ signifies the symmetric logarithmic activation function, with its inverse denoted by $ISLA(\cdot)$. These functions are detailed in Section~\ref{sec:method:Unet}. The loss function $\mathcal{L}(\theta)$ comprises two principal components. The first component calculates the mean squared difference between the predictions of the Unet, processed through the $ISLA$($f_\theta(\vec{m}_i)$), and the true target $H_{demag_i}$. The second component evaluates the mean squared difference between the direct predictions of the Unet $f_\theta(\vec{m}_i)$ and the true targets processed by the $SLA$($H_{demag_i}$), scaled by a weight factor $\lambda$.
Incorporating the weight \(\lambda\), the loss function offers a flexible strategy to balance adaptability to errors from both the $SLA(\cdot)$ and its inverse $ISLA(\cdot)$, optimizing predictive accuracy.


\paragraph{Physics-based symmetry augmentation}

To enhance model robustness, data augmentation techniques were incorporated, including rotations of 90°, 180°, and 270° counterclockwise, and flipping along both the X and Y axes. These augmentation methods accord to the physical symmetry of the demagnetizing field \( H_d \), thereby enhancing the predictive accuracy in diverse configurations.

\subsection{Evaluation with LLG dynamics}
\label{sec:emthod:vortex}

While the Unet method's quality can initially be gauged by its prediction accuracy through MSE loss on training and/or validation data, a more valuable evaluation involves scrutinizing how well the simulation outputs align between Unet/LLG and the traditional FFT/LLG methods. One approach to evaluation entails verifying whether the trajectories of LLG evolution are consistent between simulations conducted with Unet and FFT, respectively. However, quantifying the degree of similarity between two trajectories presents a complex challenge. To solve the problem, we notice that the convergence state holds greater importance than the intermediate states on the evolutionary trajectory. Therefore, our chosen method of evaluation focuses on comparing the convergence states by examining the vortices present within those states.

Magnetic vortices, encompassing both vortex and anti-vortex formations, represent stable topologies commonly found in soft magnetic films. The quantity and arrangement of vortices in a stable state heavily rely on the film's size, the material's magnetic parameters, and the initial distribution of magnetization. In our study, we assess the Unet/LLG approach's effectiveness by examining the magnetization evolution in soft magnetic films that have been randomly initialized. Precision is determined through a dual-tiered complexity measure: (i) low-level complexity, measured by the accurate prediction of the number of vortices and anti-vortices, and (ii) high-level complexity, assessed by accurately predicting both the number and specific properties of vortices, including their orientation and polarization.

\textbf{Vortex numbers:} 
Within the framework of continuous micromagnetic simulations, the identification of a single vortex or anti-vortex can be characterized by the winding number \cite{PhysRevLett.97.177202},

\begin{equation}
\label{eq:windnum}
W = \frac{1}{2\pi} \oint_{\partial S} \hat{m} \times d\hat{m} 
  = \frac{1}{2\pi} \int_{S} \frac{\partial \hat{m}}{\partial x} \times \frac{\partial \hat{m}}{\partial y} dxdy,
\end{equation}

\noindent where $S$ denotes the selected in-plane region under consideration, with $\partial S$ representing its boundary. A winding number $W = 1$ signifies the presence of a regular vortex, whereas $W = -1$ indicates an anti-vortex. Building upon this, we introduce the concept of \textbf{winding density} in the finite-differential scheme:

\begin{equation}
\begin{aligned}
    WD[i,j] &= (m_x[i+1,j] - m_x[i-1,j]) \cdot (m_y[i,j+1] - m_y[i,j-1]) / 8\pi \Delta x \Delta y \\
            &- (m_y[i+1,j] - m_y[i-1,j]) \cdot (m_x[i,j+1] - m_x[i,j-1]) / 8\pi \Delta x \Delta y .
\end{aligned}
\end{equation}

\noindent Subsequently, the summation and the absolute summation of the winding density ($WD$) are computed as below:

\begin{equation}
\begin{aligned}
    WD_{sum} = \sum_i \sum_j WD[i,j], \\
    WD_{abs} = \sum_i \sum_j abs(WD[i,j]),
\end{aligned}
\end{equation}

\noindent These calculations facilitate the identification of the total number of vortices and anti-vortices, according to the following equation:

\begin{equation}
\begin{aligned}
    N_{vortex} = WD_{abs} + WD_{sum}, \\
    N_{antiv} = WD_{abs} - WD_{sum}.
\end{aligned}
\end{equation}

We define vortex number precision as the proportion of Unet/LLG simulations that accurately predict the quantity of vortices and anti-vortices at the convergence state. This is achieved by comparing the outcomes at the convergence states between simulations conducted with Unet/LLG and the traditional FFT/LLG.

\textbf{Vortex Properties:} For each vortex, we analyze two attributes: orientation and polarization, whereas for each anti-vortex, we only consider its polarization (since anti-vortices do not possess orientation). The orientation of a vortex, whether clockwise or anti-clockwise, is determined by examining the curl of the magnetization, denoted as $\nabla \times \hat{m}$. The polarization of a vortex, being either positive or negative, is assessed by evaluating the perpendicular component $m_z$ within the vortex core region. Vortex property precision is defined as the proportion of Unet/LLG simulations resulting in convergence states where both the count and specific properties (orientation and polarization) of vortices and anti-vortices match the outcomes with FFT/LLG.

\textbf{Precision evaluation:} Each substance sample is divided into 32 $\times$ 32 blocks, with a random in-plane magnetization direction assigned to each block. Consequently, all cells within the same block are initialized with the identical spin. Initiating Unet/LLG iterations directly from this complex initial state often results in significant simulation errors. To address this issue, we implement a cooling procedure, which conducts FFT/LLG iterations until a predetermined number of vortices ($InitCore$) is obtained. It becomes evident that an increase in cooling iterations leads to fewer vortices and consequently, a more stable system. Upon achieving a relatively stable state, the Unet/LLG simulation proceeds until it reaches convergence.

Our findings indicate that with $InitCore$ set to 5, the cooling procedure constitutes approximately 2\% to 8\% of the total iterations. Notably, this approach significantly enhances the accuracy of the Unet/LLG simulation. The outcomes of employing this specific setting ($InitCore=5$) are detailed in Table~\ref{tab:vortex}, and results achieved using other $InitCore$ values can be found in the Extended Data sheet.

\textbf{Phase Diagram:} To further demonstrate the Unet/LLG approach's applicability in micromagnetic research, phase diagrams depicting the ground state of soft magnetic thin films were presented. Samples were configured into square shapes, with sizes ranging from 32 to 128 (corresponding to geometric sizes from 96 nm to 384 nm) and the default material parameters. The initialization of magnetization configurations involves dividing the sample into uniformly magnetized blocks. Each block comprises $2 \times 2 \times 2$ cells, with magnetization oriented along a randomly chosen in-plane direction. Convergent states are categorized into three types: (1) devoid of any vortices (single domain), (2) retaining a single vortex core, and (3) containing multiple vortices. For every sample size, 1000 simulations with randomly initialized samples were performed using LLG simulation to achieve a convergent state, and the probabilities with the three state type (0/1/multiple vortices) were calculated. The phase diagram results, derived from both FFT/LLG and Unet/LLG methods, are illustrated in Fig.\ref{fig:vortex_sta}(j)(k). Note that the Unet/LLG simulation employed a cooling process that achieves a vortex core count of 10 ($InitCore=10$).

\subsection{Evaluation with MH estimation}
\label{sec:method:mh}

In the field of micromagnetics, the demagnetization curve, also referred to as the MH curve or hysteresis loop, represents the relationship between a material's magnetic field (H) and its magnetization (M). This curve illustrates how a material's magnetization varies with an applied magnetic field, playing a vital role in elucidating the magnetic characteristics of materials\cite{stoner1948mechanism,zhu1988micromagnetic, victora1987quantitative,weller2016fept,nakamura2018current}. The MH curve can be approximated through simulation, in which the external field $H_{ext}$ is gradually altered from a significant positive value to a significant negative value, with the magnetization configuration being stabilized at each phase via the LLG dynamics. By considering the FFT/LLG simulation as the ground truth, the efficacy of the Unet/LLG approach is assessed by comparing the congruence of the MH curves produced by both methods.

\begin{figure}[htb!]
    \centering
    \includegraphics[width=1.0\textwidth]{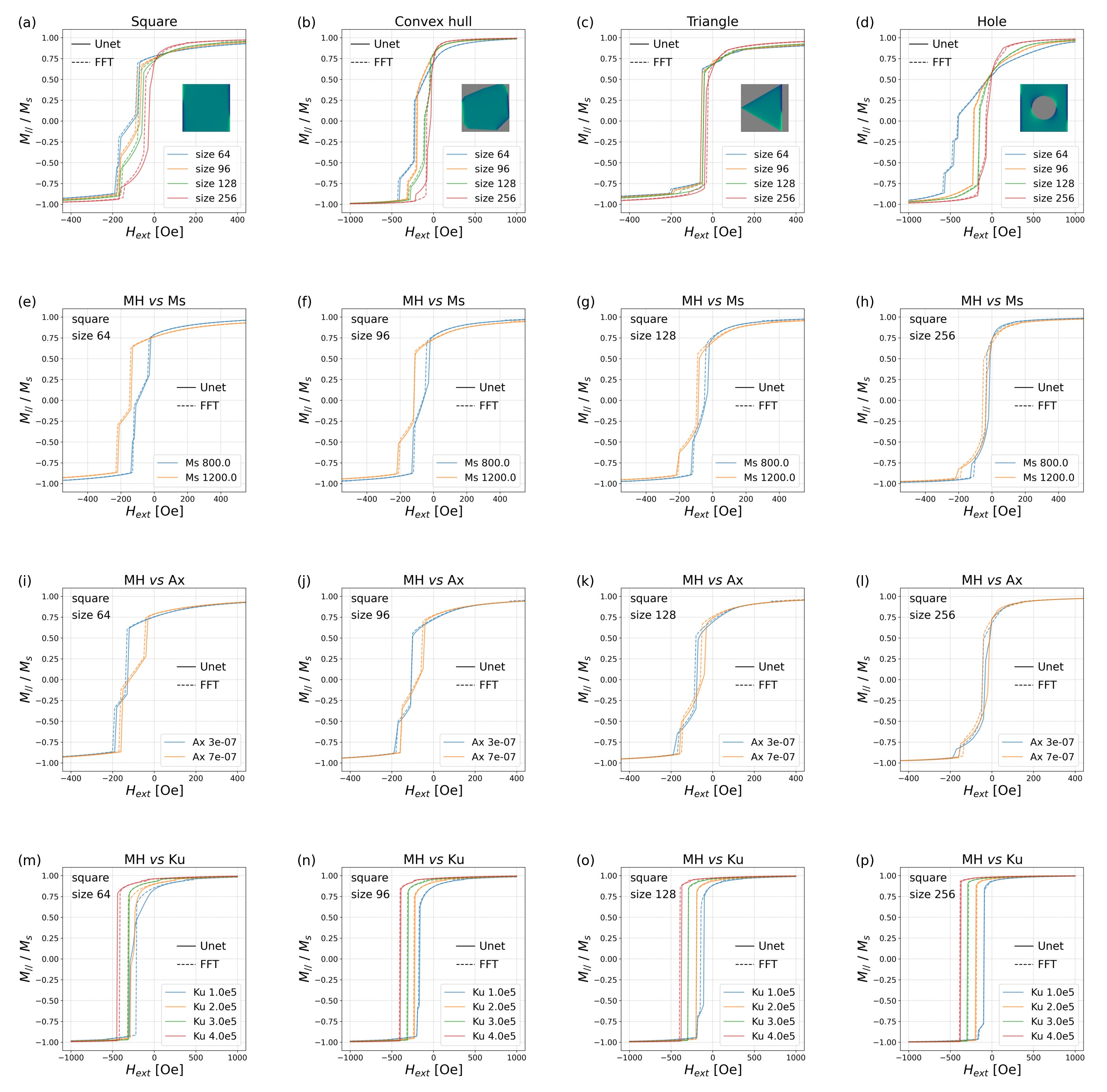}
    \caption{
    This collection showcases simulated MH curves for samples distinguished by their sizes, shapes, and unique magnetic parameters. Figures (a-d) depict simulated MH curves corresponding to samples with distinct geometries: (a) a square, (b) a convex hull, (c) a triangle, and (d) a square featuring a central hole. Figures (e-h) illustrate simulated MH curves for samples exhibiting a range of saturation magnetization ($M_s$) values. Figures (i-l) present simulated MH curves for samples with varying levels of exchange stiffness ($A_x$). Figures (m-p) exhibit simulated MH curves influenced by different uniaxial anisotropy energy densities ($K_u$). Any simulation parameters not explicitly mentioned in the figures adhere to default settings, including a square shape, saturation magnetization ($M_s$) of 1000 emu/cc, exchange stiffness ($A_x$) of $0.5\times10^{-6}$ erg/cm, and uniaxial anisotropy energy density ($K_u$) of 0 erg/cc. The applied external magnetic field ($H_{ext}$) and the easy axis associated with $K_u$ are both oriented along the $x$ direction.
    }
    \label{fig:MHall}
\end{figure}

In our experiment, the simulation was conducted across a spectrum of sample sizes, shapes, and magnetic material parameters. During the simulation, an external magnetic field \(H_{ext}\) was applied in the \(x\) direction, initiating at 1000 Oe and incrementally decreasing to -1000 Oe in steps of 10 Oe. At every specified magnetic field value, the system was allowed to reach a stabilized state prior to any subsequent change in the field. The outcomes of these simulations are presented in Fig.~\ref{fig:MH} and Fig.~\ref{fig:MHall}.

\subsection{Speed Acceleration with TensorRT}
\label{sec:method:speed}

Unet/LLG is a deep learning approach based on neural network model implemented with the PyTorch package. This allows it to leverage advanced optimization techniques specifically designed for deep neural nets, such as pruning, quantization, knowledge distillation. These techniques often speed up computation significantly. 

In this study, we use TensorRT to improve the inference speed of the Unet model. TensorRT, renowned for its capability to optimize deep neural nets, ensures low latency and high throughput, crucial for real-time applications. The optimization methods include layer fusion, kernel auto-tuning, dynamic tensor memory\cite{prasanna2019deep}. These methods enable more efficient execution on GPU platforms.

To compile our Unet model with TensorRT, we input the trained Unet model to torch\_tensorrt and set the precision to be float16. An optimized TensorRT model is then returned and used to perform efficient inference. 

\backmatter

\section*{Data availability}
\label{sec:Data}
The datasets used in our study are generated by a script, which is publicly available for full reproducibility of our results. Researcher can access the script at \url{https://github.com/Caiyq2019/NeuralMAG/tree/main/utils}.

\section*{Code availability}
\label{sec:Code}
The software tool developed for this research has been packaged and made accessible to researchers. Additionally, scripts to replicate the experimental results presented in this paper are included. Access is provided via the following GitHub repository: \url{https://github.com/Caiyq2019/NeuralMAG/tree/main/}.

\section*{Declarations}

The authors declare no competing interests. 

\section*{Preprint Notice}
This preprint has not undergone peer review (when applicable) or any post-submission improvements or corrections. The Version of Record of this article is published in [insert journal title], and is available online at \url{https://doi.org/[insert DOI]}.

\section*{Author information}

\paragraph{Contributions}
Y.Q. Cai and J.N. Li initiated the project, with leadership by D. Wang. They collectively formulated the integration of deep learning with micromagnetic simulations. The design of the deep learning framework was a collaborative effort between Y.Q. Cai and D. Wang, whereas J.N. Li focused on the development of accuracy evaluation methods for micromagnetic simulations. Y.Q. Cai led the training and testing of the deep learning models, and alongside J.N. Li, conducted NeuralMAG experiments. D. Wang contributed theoretical insights into hierarchical computations and complexity. All authors designed figures and co-wrote the manuscript, contributing to revisions and finalizing the paper from diverse perspectives.

\paragraph{Corresponding author}
Correspondence to J.N. Li and D. Wang


\bibliography{sn-bibliography}

\newpage
\section*{Extended Data}

\begin{table}[htb!]
\centering
\caption{
Accuracy of magnetic ground state predictions using Unet/LLG simulations, which commence from varied starting states achieved through a cooling procedure. This procedure utilizes FFT/LLG iterations to reach a state with a pre-defined number of vortices, termed InitCore. The term 'Cooling Proportion' quantifies the fraction of the entire simulation process, measured in iterations, that the cooling procedure constitutes. The samples utilized in this study are of a square shape, with all material settings held constant. The training samples are available in two sizes: 32 and 64.
}

\label{tab:ext:size}
\begin{tabular}{@{}ccm{2cm}m{2cm}m{2cm}@{}}
\toprule
\textbf{InitCore} & \textbf{\begin{tabular}[c]{@{}c@{}}Cooling \\ Proportion\end{tabular}} & \textbf{Sample Size} & \textbf{\begin{tabular}[c]{@{}c@{}}Vortex Number \\ Precision\end{tabular}} & \textbf{\begin{tabular}[c]{@{}c@{}}Vortex Property \\ Precision\end{tabular}} \\
\midrule
5  & 2.80\% & \multirow{3}{*}{32}  & 97.00\% & 94.00\% \\
10 & 1.60\% &                      & 95.00\% & 90.00\% \\
20 & 0.80\% &                      & 90.00\% & 80.00\% \\
\hline
5  & 4.70\% & \multirow{3}{*}{64}  & 97.00\% & 97.00\% \\
10 & 2.80\% &                      & 97.00\% & 94.00\% \\
20 & 1.70\% &                      & 94.00\% & 90.00\% \\
\hline
5  & 8.00\% & \multirow{3}{*}{128} & 97.00\% & 93.00\% \\
10 & 4.10\% &                      & 94.00\% & 90.00\% \\
20 & 2.40\% &                      & 91.00\% & 85.00\% \\
\bottomrule
\end{tabular}
\end{table}

\begin{table}[htb!]
\centering
\caption{
Accuracy of magnetic ground state predictions using the Unet/LLG simulation, initiated from various starting states. Notations and experimental settings align with those detailed in Table~\ref{tab:ext:size}, with the sole exception being that the sample shapes are randomly generated.
}

\label{tab:ext:shape}
\begin{tabular}{@{}ccm{2cm}m{2cm}m{2cm}@{}}
\toprule
\textbf{InitCore} & \textbf{\begin{tabular}[c]{@{}c@{}}Cooling \\ Proportion\end{tabular}} & \textbf{Sample Size} & \textbf{\begin{tabular}[c]{@{}c@{}}Vortex Number \\ Precision\end{tabular}} & \textbf{\begin{tabular}[c]{@{}c@{}}Vortex Property \\ Precision\end{tabular}} \\
\midrule
5  & 2.80\% & \multirow{3}{*}{32}  & 98.00\% & 93.00\% \\
10 & 1.40\% &                      & 94.00\% & 80.00\% \\
20 & 0.70\% &                      & 85.00\% & 70.00\% \\
\hline
5  & 4.20\% & \multirow{3}{*}{64}  & 99.00\% & 97.00\% \\
10 & 2.40\% &                      & 99.00\% & 96.00\% \\
20 & 1.50\% &                      & 100.00\% & 93.00\% \\
\hline
5  & 7.20\% & \multirow{3}{*}{128} & 89.00\% & 86.00\% \\
10 & 4.00\% &                      & 87.00\% & 82.00\% \\
20 & 2.30\% &                      & 80.00\% & 70.00\% \\
\bottomrule
\end{tabular}
\end{table}

\begin{table}[htb!]
\centering
\caption{
Accuracy of magnetic ground state predictions using the Unet/LLG simulation, initiated from various starting states. Notations and experimental settings align with those detailed in Table~\ref{tab:ext:size}, with the sole exception being that the material parameters are randomly assigned.
}

\label{tab:prediction_accuracy_randomset}
\begin{tabular}{@{}ccm{2cm}m{2cm}m{2cm}@{}}
\toprule
\textbf{InitCore} & \textbf{\begin{tabular}[c]{@{}c@{}}Cooling\\ Proportion\end{tabular}} & \textbf{Sample Size} & \textbf{\begin{tabular}[c]{@{}c@{}}Vortex Number \\ Precision\end{tabular}} & \textbf{\begin{tabular}[c]{@{}c@{}}Vortex Property \\ Precision\end{tabular}} \\
\midrule
5  & 1.90\% & \multirow{3}{*}{32}  & 98.00\% & 98.00\% \\
10 & 1.00\% &                      & 95.00\% & 92.00\% \\
20 & 0.60\% &                      & 87.00\% & 73.00\% \\
\hline
5  & 3.90\% & \multirow{3}{*}{64}  & 98.00\% & 98.00\% \\
10 & 2.20\% &                      & 98.00\% & 96.00\% \\
20 & 1.30\% &                      & 96.00\% & 92.00\% \\
\hline
5  & 6.50\% & \multirow{3}{*}{128} & 92.00\% & 91.00\% \\
10 & 3.30\% &                      & 84.00\% & 79.00\% \\
20 & 1.80\% &                      & 81.00\% & 75.00\% \\
\bottomrule
\end{tabular}
\end{table}

\begin{table}[htb!]
\centering
\caption{
Comparative analysis of computational speed and GPU memory usage: evaluating FFT/LLG, Unet/LLG, and TensorRT-accelerated Unet/LLG simulations on an Nvidia RTX-3090 GPU.
}

\label{tab:speed_all}
\begin{tabular}{@{}c c c c c c c@{}}
\toprule
\multirow{2}{*}{\textbf{size}} &
  \multicolumn{2}{c}{\textbf{FFT}} &
  \multicolumn{2}{c}{\textbf{Unet}} &
  \multicolumn{2}{c}{\textbf{Unet-TensorRT}} \\
\cmidrule(lr){2-3} \cmidrule(lr){4-5} \cmidrule(lr){6-7}
 &
  \textbf{\begin{tabular}[c]{@{}c@{}}Iteration \\ time\end{tabular}} &
  \textbf{\begin{tabular}[c]{@{}c@{}}$H_{demag}$ time\\ per iteration\end{tabular}} &
  \textbf{\begin{tabular}[c]{@{}c@{}}Iteration \\ time\end{tabular}} &
  \textbf{\begin{tabular}[c]{@{}c@{}}$H_{demag}$ time\\ per iteration\end{tabular}} &
  \textbf{\begin{tabular}[c]{@{}c@{}}Iteration \\ time\end{tabular}} &
  \textbf{\begin{tabular}[c]{@{}c@{}}$H_{demag}$ time\\ per iteration\end{tabular}} \\
\midrule
32   & 5.80E-03 s  & 2.28E-03 s & 1.10E-02 s & 8.40E-03 s & 5.40E-03 s & 1.56E-03 s \\
128  & 5.90E-03 s  & 2.36E-03 s & 1.20E-02 s & 8.00E-03 s & 5.50E-03 s & 1.60E-03 s \\
512  & 2.00E-02 s  & 1.32E-02 s & 1.90E-02 s & 1.20E-02 s & 1.00E-02 s & 3.40E-03 s \\
1024 & 8.10E-02 s  & 5.60E-02 s & 6.70E-02 s & 4.00E-02 s & 3.70E-02 s & 1.04E-02 s \\
2048 & 3.60E-01 s  & 2.56E-01 s & 2.60E-01 s & 1.60E-01 s & 1.50E-01 s & 3.80E-02 s \\
3072 & --          & --         & 6.00E-01 s & 3.64E-01 s & 3.30E-01 s & 8.40E-02 s \\
\bottomrule
\end{tabular}
\end{table}

\end{document}